%% file: main.tex
\documentclass[final]{neus2025}

\usepackage{longtable}
\usepackage{booktabs}
\usepackage{latexsym}
\usepackage{hyperref}
\usepackage{cite}

\usepackage{multirow}
\usepackage{multicol}
\usepackage{mdframed}

\usepackage[T1]{fontenc}
\usepackage{microtype}

\usepackage{times}

\author{%
 \Name{Peiyang Song} \Email{psong@caltech.edu}\\
 \addr California Institute of Technology, Pasadena, CA, U.S.A.
 \AND
 \Name{Kaiyu Yang} \Email{kaiyuy@caltech.edu}\\
 \addr California Institute of Technology, Pasadena, CA, U.S.A.
 \AND
 \Name{Anima Anandkumar} \Email{anima@caltech.edu}\\
 \addr California Institute of Technology, Pasadena, CA, U.S.A.
}

\newcommand{\lc}{{Lean Copilot}}

\title[Lean Copilot]{Lean Copilot: Large Language Models as Copilots\\for Theorem Proving in Lean}

\begin{document}

\maketitle

\input{sections/abstract}


\input{sections/intro}
\input{sections/related}
\input{sections/leancopilot}
\input{sections/tools}
\input{sections/experiments}
\input{sections/conclusion}

\acks{Peiyang Song was supported by the Summer Undergraduate Research Fellowship at Caltech. Kaiyu Yang was supported by the Computing, Data, and Society Postdoctoral Fellowship at Caltech. Anima Anandkumar is partially supported by the Bren endowed chair and Schmidt Sciences AI2050 senior fellowship. We appreciate the valuable feedback from members of the Anima AI+Science Lab on an initial version of this paper, from the Lean community on the releases of our tool, and from the anonymous reviewers at NeuS 2025. We thank Scott Morrison for suggestions on simplifying Lean Copilot’s installation and Mac Malone for helping implement it. We thank Jannis Limperg for supporting our LLM-generated tactics in Aesop.}

\bibliography{ref}

\appendix
\input{sections/artifacts}
\input{sections/generalize}
\input{sections/detailed}

\input{sections/impact}

\end{document}

%% file: sections/abstract.tex
\begin{abstract}

Neural theorem proving combines large language models (LLMs) with proof assistants such as Lean, where the correctness of formal proofs can be rigorously verified, leaving no room for hallucination.
With existing neural theorem provers pretrained on a fixed collection of data and offering valuable suggestions at times, it is challenging for them to continually prove novel theorems in a fully autonomous mode, where human insights may be critical.
In this paper, we explore LLMs as copilots that assist humans in proving theorems. 
We introduce {\lc}, a general framework for running LLM inference natively in Lean. 
It enables programmers to build various LLM-based proof automation tools that integrate seamlessly into the workflow of Lean users.
Lean users can use our pretrained models or bring their own ones that run either locally (with or without GPUs) or on the cloud. 
Using {\lc}, we build LLM-based tools that suggest proof steps, complete proof goals, and select relevant premises. 
Experimental results on the \textit{Mathematics in Lean} textbook demonstrate the effectiveness of our method compared to existing rule-based proof automation in Lean (\textsc{aesop}), confirming the significance of having LLMs integrated into the theorem proving workflow in Lean.
When assisting humans, {\lc} requires only \textbf{2.08} manually-entered proof steps on average (3.86 required by \textsc{aesop}); when automating the theorem proving process, {\lc} automates \textbf{74.2\%} proof steps on average, \textbf{85\%} better than \textsc{aesop} (40.1\%). 
We open source all code and artifacts under a permissive MIT license to facilitate further research.

\end{abstract}

%% file: sections/intro.tex
\section{Introduction}
\label{sec:intro}

\begin{figure*}[t]
  \centering
  \vspace{-8mm}
    \includegraphics[width=1.0\linewidth]{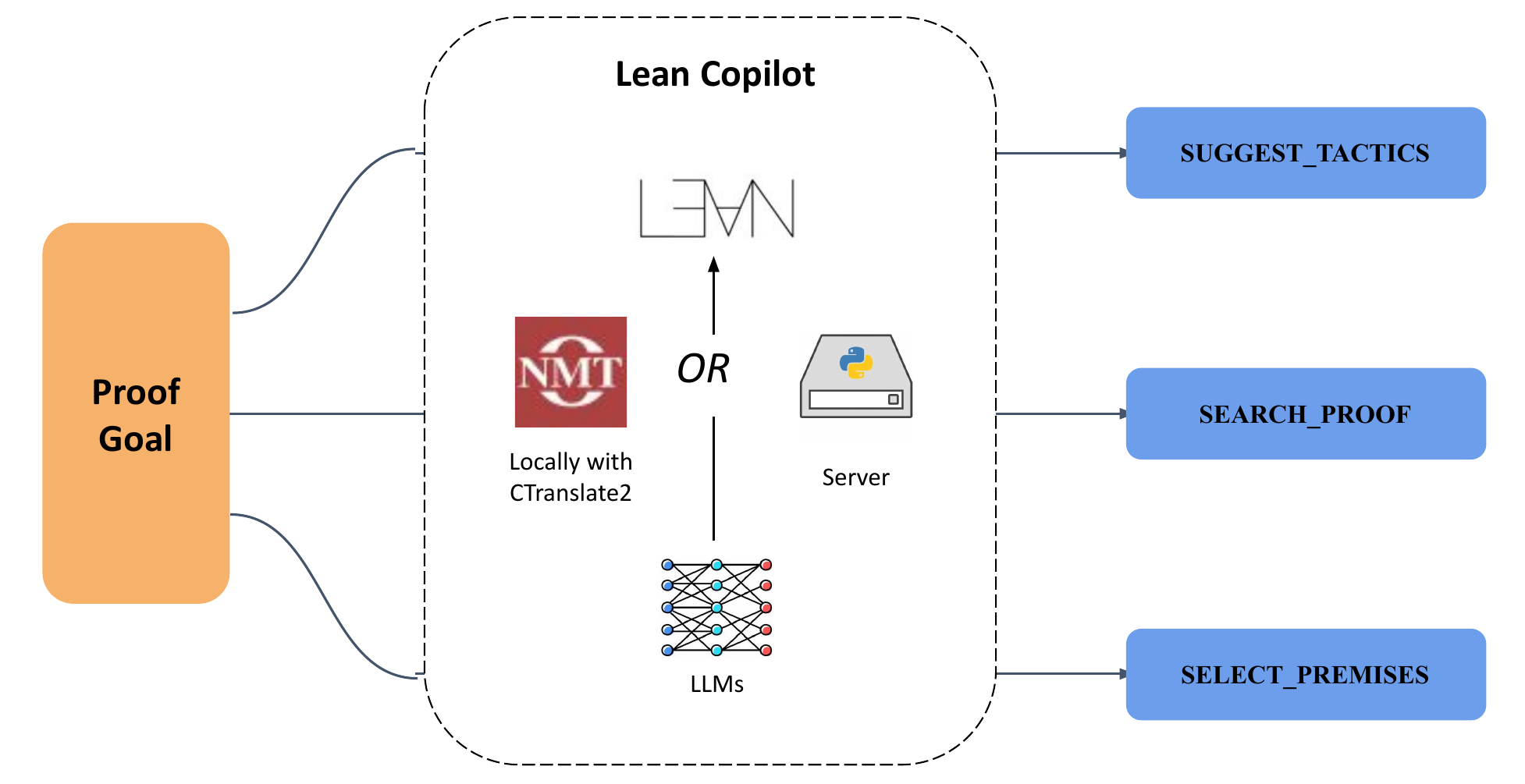}
  \vspace{-6mm}
  \caption{{\lc} provides a general framework for running LLM inference in Lean, either locally via CTranslate2 or on a server. This framework enables creating various proof automation tools, including those for tactic suggestion, proof search, and premise selection.}
  \vspace{-8mm}
  \label{fig:lean_copilot}
\end{figure*}

Neural theorem proving \citep{li2024surveydeeplearningtheorem} combines the strong learning capability of \textit{neural} networks with the rigor of \textit{symbolic} proof checking. 
The \textit{neural} component trains LLMs to effectively generate formal proofs, yet LLMs' tendency to hallucinate \citep{hallucination} prevents the generated proofs from guaranteed correct.
The \textit{symbolic} component uses proof assistants to verify proof correctness, but the interactive nature of proof assistants require substantial amount of human efforts to formalize proofs.
Note that each component's weakness is naturally complemented by the other.
Combining both, neural theorem proving trains LLMs to generate candidate proofs and verifies them by proof assistants, forming a powerful neuro-symbolic system for formal reasoning \citep{yang2024formalmathematicalreasoningnew}.

On the symbolic side, Lean \citep{de2015lean} has been a widely used proof assistant especially in mathematical theorem proving, thanks to its large math library \textit{Mathlib4} \citep{mathlib} with over 189K theorems\footnote{Statistics obtained from \url{leanprover-community.github.io/mathlib_stats.html}, Feb. 24, 2025.} from diverse domains, which is still expanding. 
On the neural side, LeanDojo \citep{yang2023leandojo} novelly augments tactic generation with premise retrieval, and pairs tactic (individual proof step in Lean) generation with proof search to form complete proofs in a GPT-\textit{f} \citep{leangptf} style.
Later works propose methods to further improve both tactic generation \citep{lin2024leanstarlearninginterleavethinking} and proof search \citep{huang2025leanprogressguidingsearchneural}.
An alternative approach is whole-proof generation \citep{first2023baldur}, where synthetic data is used to address the data scarcity problem \citep{frieder2024datamathematicalcopilotsbetter}.
By scaling up, this approach has given birth to powerful proof-generation LLMs, marked by DeepSeek-Prover-v1.5 \citep{xin2024deepseekproverv15harnessingproofassistant} and Goedel-Prover \citep{lin2025goedelproverfrontiermodelopensource}.

All these existing LLM-based provers aim to prove theorems fully autonomously without human intervention. 
They wrap the proof assistant (e.g., Lean) into a gym-like {\citep{brockman2016openai}} environment.
The model interacts with the proof environment and is evaluated by the number of test theorems it proves. 
The interaction happens solely on the backend server, without any collaboration with humans. 
While an autonomous AI mathematician is desirable in the long run, current LLMs often fail to prove theorems that are truly novel or challenging, especially those from a different domain than the training data \citep{zheng2022minif2f}. 
This is in part because each branch of mathematics uses different lemmas and requires distinct intuitions, limiting the generalization of LLMs. 
However, with the development in Lean mostly focused on Mathlib4 and other large-scale formalization projects \citep{gowers2023conjecturemarton, liquid2022}, proving in new domains is inevitably important. 

In practice, proving theorems in Lean requires a mix of mathematical intuition and tedious labor such as looking up premise names from the large math library. 
For mathematicians, the latter is particularly time-consuming.
While current LLMs fall short in the former on novel theorems, we hypothesize that these LLMs are able to effectively aid the latter and save human labor.
AI should act as a copilot in theorem proving: an assistant that eases routine proof construction while enabling experts to guide the overall process.

\paragraph{{\lc}.}

A seamless collaboration between mathematicians and LLMs would require running inference of the pretrained LLMs natively in Lean, which is however a symbolic system not designed for neural network applications. 
To bridge this gap, we present {\lc}, a neuro-symbolic framework for developing LLM-based proof automation in Lean.
It addresses a core technical challenge: \emph{running LLM inference in Lean} (Section~\ref{sec:leancopilot}).

{\lc} provides two levels of functionalities. 
First, for Lean users, we build three LLM-based tools to assist theorem proving (Section~\ref{sec:tools}). 
They offer suggestions for the next tactic (\textsc{suggest\_tactics}), search for a complete verified proof (\textsc{search\_proofs}), and select the most relevant premises to apply next (\textsc{select\_premises}).
Second, for developers to build other LLM-based tools in Lean, we offer two low-level interfaces that run text-to-text and text-to-vector generations (Appendix~\ref{sec:generalize}).
Under the hood, we support running LLM inference either locally or via a server process.
We use ReProver \citep{yang2023leandojo} as our default model due to its unique capability in premise retrieval, while we support users bringing their own models easily to {\lc} as well.

We evaluate (Section~\ref{sec:experiments} \& Appendix~\ref{sec:detailed}) our tools on the whole \textit{Mathematics in Lean} textbook \citep{avigad2020mathematics}.
\textsc{suggest\_tactics} and \textsc{search\_proof} represent two levels of automation.
\textsc{suggest\_tactics} generates the next \textit{one} tactic based on a proof goal.
\textsc{search\_proof} further combines single tactic generation with a proof search algorithm (by default a best-first search with LLM confidence score as critic), and generates a complete proof.
\textsc{aesop} automates on the same level by searching for a complete proof, but its tactics are from a pre-defined list rather than generated by LLMs.
Results show that \textsc{search\_proofs} alone (without humans) can automate \textbf{74.2\%} proof steps on average, a performance \textbf{85\%} higher than the original \textsc{aesop} and \textbf{27\%} higher than our tactic generation tool \textsc{suggest\_tactics}. 
It can also better assist humans than the two baselines, requiring fewer tactics to be entered manually by humans (only \textbf{2.08} compared to 3.86 required by \textsc{aesop}).
Its significantly better performance than \textsc{aesop} confirms the benefit of integrating LLMs into the theorem proving pipeline in Lean, and its enhancement over single tactic generation shows the usefulness of adding a proof search algorithm on the top.

{\lc} with our automation tools can be installed easily as a Lean package and run on most hardware, paving the way for wide adoption and impact on the Lean community (Appendix~\ref{sec:impacts}). 
They are among the first steps in making LLMs accessible to human users of proof assistants, which we hope will initiate a positive feedback loop where proof automation leads to better data and ultimately improves LLMs on mathema. 
We open source all code on Github (Appendix~\ref{sec:artifacts}) at \url{https://github.com/lean-dojo/LeanCopilot} to facilitate future research. 
We make all evaluation code and detailed results public at \url{https://github.com/Peiyang-Song/mathematics_in_lean/tree/full-scale-experiment} for full transparency.

%% file: sections/related.tex
\section{Related Work \& Preliminaries}
\label{sec:related}

\paragraph{Neural Theorem Proving.}

Neural networks have been used to prove formal theorems by interacting with proof assistants. 
They can select premises \citep{irving2016deepmath, wang2017premise, goertzel2022isabelleenigma, mikula2023magnushammer}, suggest tactics (proof steps) \citep{whalen2016holophrasmneuralautomatedtheorem, huanggamepad, li2021isarstep, kaliszykholstep}, and generate complete proofs \citep{first2023baldur, wang2024provingtheoremsrecursively, xin2024deepseekproverv15harnessingproofassistant, lin2025goedelproverfrontiermodelopensource}.
They can also aid in auxiliary tasks that help the development of neural theorem proving, such as conjecturing \citep{dong2025stpselfplayllmtheorem, bengio2024machinelearninginformationtheory, johansson2023exploring, poesia2024learningformalmathematicsintrinsic} and autoformalization \citep{wu2022autoformalization, azerbayev2023proofnet, jiang2023multilingualmathematicalautoformalization, ying2024leanworkbooklargescalelean}.
Early works on neural theorem proving often use graph neural networks \citep{yang2019learning,bansal2019holist,bansal2019learning,paliwal2020graph,wang2020learning,first2020tactok,rabe2021mathematical,sanchezstern2023passport}, 
whereas more recent works focus on Transformer-based \citep{vaswani2017attention} language models \citep{polu2020generative,jiang2021lisa,zheng2022minif2f,han2022proof,lample2022hypertree,jiang2022thor,Liu2023fimo,polu2023formal,wang2023dt,first2023baldur,yang2023leandojo, wang2023legoprover}. 
While these works have demonstrated the capability of LLMs in theorem proving, none of them has led to practical and open-source tools enabling LLMs to be used \textit{directly} in proof assistants.

\paragraph{Automation within Proof Assistants.}

Proof automation has been studied extensively using formal methods. 
Many efficient decision procedures are available for specific domains, such as satisfiability modulo theories \citep{ekici2017smtcoq, martínez2019metafproofautomationsmt}, linear arithmetic \citep{besson2007fast}, and commutative rings \citep{gregoire2005proving}. 
Proof automation tools in Lean are usually wrapped up in individual tactics for a seamless integration into the normal workflow.
Lean's \texttt{apply?} tactic tries to find premises that unify symbolically with the current goal. 
There are also general-purpose proof search tactics such as \textsc{aesop} \citep{limperg2023aesop} in Lean and \textsc{auto} in Coq. 
They search for proofs by combining a set of rules with algorithms such as best-first search. 
The rules are configured manually by users and fixed throughout the proof search, instead of being selected or ranked based on remaining proof goals.

Many classical machine learning algorithms have been used for proof automation. 
Hammers \citep{blanchette2016hammering, bohme2010sledgehammer, czajka2018hammer} often use machine learning for premise selection, 
and then outsource the proof goal and the selected premises to external automated theorem provers in first-order logic. 
TacticToe \citep{gauthier2021tactictoe} and Tactician \citep{blaauwbroek2020tactician} predict tactics using the k-nearest neighbors algorithm (KNN) with handcrafted features.
\citet{piotrowski2023machinelearned} and Geesing \citep{geesing2023premise} have implemented Naive Bayes, random forests, and kNN within Lean for premise selection. 
There have been prior and concurrent efforts exploring using neural networks or specifically LLMs in proof assistants \citep{coqsynthesis, leangptf, llmstep, sagredo}. 
All of them run models in Python and make requests to the models from the proof assistant. 
In contrast, we support running LLMs natively in Lean (details in Section~\ref{sec:leancopilot}).

\paragraph{Human-AI collaboration in Formal Mathematics.}

\citet{collins2023evaluating} has investigated using LLMs to assist human mathematicians by holding conversations in natural language.
In formal math, \citet{doi:10.1073/pnas.2318124121} study the patterns of human-AI interaction for theorem proving in Isabelle, by building an external platform.
The paradigm of using LLMs as ``copilots'' to assist humans originates from software programming, as witnessed by the huge success of tools like GitHub Copilot \citep{chen2021evaluating, yetiştiren2023evaluatingcodequalityaiassisted}. 
Such interactions usually require the AI tools to be incorporated deeply into the same environment where humans are working, so that direct interactions and collaborations can be possible. 
For neural theorem proving, this means having AI tools natively in the proof assistant environment. 
To the best of our knowledge, we are the first to address this challenge by building a general framework that can run LLM inference natively in Lean.

%% file: sections/leancopilot.tex
\section{{\lc}: A General Framework for Running LLM Inference in Lean}
\label{sec:leancopilot}

For LLMs to assist humans in Lean, {\lc} provides a general framework for \emph{running LLM inference in Lean}, which Lean programmers can use to build various LLM-based applications. 
Lean is usually fast in offering environment feedback. 
This enables Lean users to reason coherently without being interrupted by a long waiting time. 
{\lc} also needs to satisfy this requirement.
Additionally, theorem proving in Lean can work well on a user's local laptop without GPUs. 
{\lc} needs to avoid adding extra hardware constraints, so it should be able to run efficiently on most hardware even without GPUs.

With the desire for fast inference and low compute requirement, since most mainstream efficient deep learning frameworks are in Python \citep{wolf2020huggingfaces, tensorflow2015-whitepaper, paszke2019pytorch, chollet2015keras}, a natural solution is to host the model in Python and make requests to it from Lean \citep{llmstep}. 
However, this approach suffers from the overhead of inter-process communication, and it requires users to perform additional setup steps that do not fit naturally into Lean's conventional workflow. 
To overcome these issues, {\lc} runs LLMs \textit{natively} in Lean through its \textit{foreign function interface (FFI)}.
All automation tools offered by {\lc} take a tactic form and can be applied just as any other regular tactic in Lean, with {\lc} itself wrapped up as a Lean package.
This makes the setup and usage seamlessly integrated into Lean's workflow.

\paragraph{Running LLMs in Lean through FFI.}
FFI is a mechanism for programs in one language to call subroutines from another language. 
Lean is partially implemented in C++ and interoperates efficiently with C++. 
Programmers can thus declare a function in Lean but implement its body in C++.
The implementation is compiled into a shared library and linked to Lean dynamically.

By default, we adopt the pretrained ReProver model from LeanDojo \citep{yang2023leandojo} for its multi-faceted capability in not only tactic generation but also premise retrieval.
ReProver is based on an encoder-decoder Transformer, ByT5 {\citep{xue2022byt5}}, that maps an input string to an output string. 
{\lc} makes the model runnable in Lean by wrapping its inference into a C++ function operating on strings, which can be called in Lean through FFI. 
As in Fig.~\ref{fig:lean_copilot}, we can run the model either locally via CTranslate2 or on a server. 
We use beam search for decoding the output sequence, with configurable hyperparameters such as temperature and the number of desired output sequences. 
This allows for multiple different output sequences being generated. 
We do not perform tokenization since ByT5 is a tokenizer-free model that works directly on UTF-8 bytes, so is ReProver.

ReProver additionally uses retrieval to select premises based on proof goals.
It conditions on the concatenation of the retrieved premises and the proof goals to generate tactics.
Its retriever is based on Dense Passage Retrieval \citep{karpukhin2020densepassageretrievalopendomain}.
With the pre-computed premise embedding from ReProver, we encode the proof goals using ReProver's encoder, and perform one forward pass with the embedding.
The encoder is part of the ReProver model and can be run through FFI functions.
Similarly, we run the retriever natively in Lean by wrapping its forward pass into a C++ function that takes an encoded state vector as input and outputs a vector of relevancy scores.
The function processes the relevancy scores to obtain a number of highest-ranked premises and returns them as a list of strings.
It can then be called in Lean through FFI.

Despite the default setup, {\lc} is a \textit{general framework} that supports \textit{user-brought models}. 
In principle, {\lc} is able to run any model with minimal changes required.
Users just need to wrap it properly to expose the APIs necessary for generation and/or encoding.

%% file: sections/tools.tex
\section{Building LLM-based Proof Automation with {\lc}}
\label{sec:tools}

The general framework for running LLM inference in Lean (Section~\ref{sec:leancopilot}) enables building various LLM-based tools for proof automation. 
In particular, we build LLM-based tools that automates three important tasks in theorem proving: tactic suggestion (Section~\ref{subsec:tactic_suggestion}), proof search (Section~\ref{subsec:proof_search}), and premise selection (Section~\ref{subsec:premise_selection}).
We also generalize {\lc} with low-level interfaces, one for text-to-text generation and the other for text-to-vector encoding (details in Appendix~\ref{sec:generalize}). 
Developers can use those interfaces to build other proof automation tools, or bring their own models to {\lc}.
We hope this feature encourages developers to build more proof automation in Lean.

\subsection{Generating Tactic Suggestions}
\label{subsec:tactic_suggestion}

When humans prove theorems in Lean, they inspect the remaining proof goals to decide for the next tactic (i.e. proof step in Lean). 
Tactics do not come from a predefined list; they are similar to programs in a domain-specific language (DSL). 
They can take arbitrary Lean terms (or take none) as parameters, and simpler tactics can be combined into compound ones. 
Users can also extend existing tactics by defining customized tactics. 
Due to these complexities, producing the right tactic can be challenging even for experienced Lean users. 

\begin{figure*}[t]
  \centering
  \vspace{-8mm}
    \includegraphics[width=1.0\linewidth]{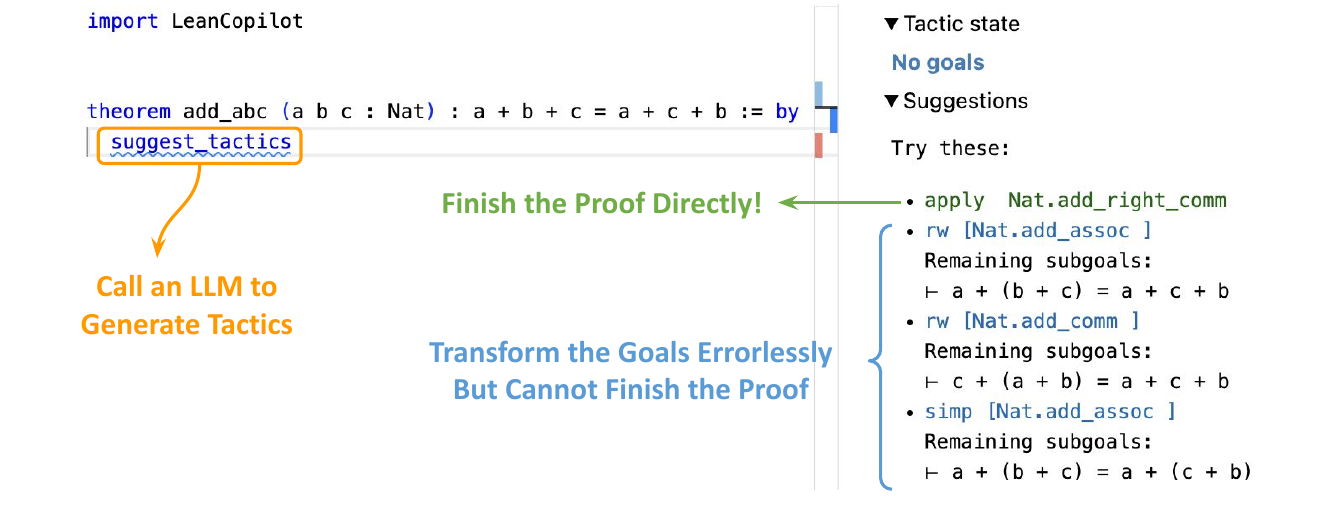}
  \vspace{-6mm}
  \caption{The frontend of {\lc}'s \textsc{suggest\_tactics}. The user imports {\lc} just as a regular Lean package, and uses \textsc{suggest\_tactics} in a proof, which calls an LLM to generate tactic candidates. After filtering, tactics that transform the proof goals errorlessly (i.e. ``tactic states'' in Lean) are shown in the InfoView (the view on the right, a vital part in Lean's workflow). The one that finishes the proof alone is colored green; others shown in blue with their respective remaining goals.}
  \vspace{-8mm}
  \label{fig:suggest_tactics}
\end{figure*}

\paragraph{Tactic Suggestion.}

We use {\lc} to build \textsc{suggest\_tactics}: a tool using LLMs to generate tactic suggestions. 
\textsc{suggest\_tactics} itself is also a tactic. 
When applied, it feeds the current proof goals into an LLM and obtains a list of tactic candidates. 
Instead of blindly returning all of them to users, we further check each tactic candidate via a meta-program in Lean that simulates running the tactic candidate on the current proof goals.
There are three possible results.
If the tactic candidate leads to errors, we drop them from the list.
Otherwise, we display them in Lean's InfoView, and color them in green or blue based on whether the tactic alone can finish the whole proof or not.

As a result, only tactics that lead to no errors are shown to the users.
We show them together with the remaining proof goals if they were applied.
This is especially useful when none of the suggested tactics directly finish a complicated proof.
Users can then use the remaining goals to choose tactics that simplify the proof goals in desired directions.
Our frontend for displaying suggested tactics in Lean's InfoView is based on Batteries\footnote{Lean 4 Batteries: \url{https://github.com/leanprover-community/batteries}.}, the standard library in Lean.
A view of the frontend is in Figure~\ref{fig:suggest_tactics}.
Users can accept a suggested tactic simply by clicking on it. 
The accepted tactic will automatically replace the \textsc{suggest\_tactics} tactic and appear at the current position in the proof.

\subsection{Searching for Complete Proofs}
\label{subsec:proof_search}

While suggesting the next proof step can be helpful, Lean proofs often consist of multiple tactics, and writing them involves trial and error. 
Neither humans nor machines can consistently produce the right tactic, so they have to backtrack and explore alternatives -- a process called \emph{proof search}. 
To address this need, we combine \textsc{suggest\_tactics} with \textsc{aesop} {\citep{limperg2023aesop}}, a rule-based, white-box proof search tool in Lean.
We empower it with LLM-generated tactics and build LLM-based proof search, wrapped up in a Lean tactic \textsc{search\_proof}. 

\paragraph{Aesop.}

\textsc{Aesop} implements best-first search and wraps it up in a Lean tactic \textsc{aesop}.
It allows users to configure how the search tree gets expanded. 
The search tree consists of proof goals as nodes. 
Initially, it has only the original goal as the root node. 
At each step, \textsc{aesop} picks the most promising unexpanded node, expands it by applying tactics from a pre-defined set, and adds the resulting nodes as its children. 
The proof search succeeds when \textsc{aesop} finds a path from the root to some goals that can be solved trivially. 
It may fail because of timeout or when \textsc{aesop} has run out of tactics to try. 

In \textsc{aesop}, tactics for expanding nodes are drawn from a set called \textit{the rule set}. 
It is configurable by users before proof search but fixed during the search, i.e., the same rule set is used for expanding all nodes, regardless of the proof goal. 
Its performance then depends critically on whether the user has configured an effective rule set, which is however often problem-dependent. 
Therefore, \textsc{aesop} cannot adaptively decide what tactics to try given intermediate proof goals during the search process. 

\begin{figure*}[t]
  \centering
  \vspace{-8mm}
    \includegraphics[width=1.0\linewidth]{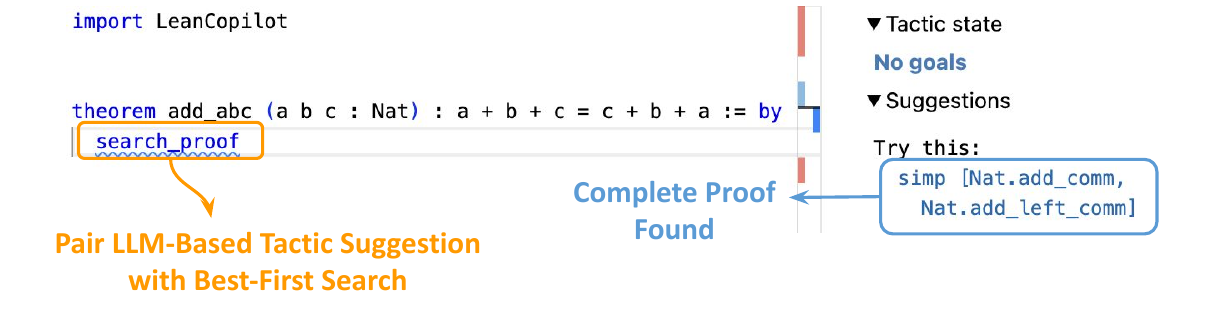}
  \vspace{-6mm}
  \caption{The frontend of {\lc}'s \textsc{search\_proof}. When \textsc{search\_proof} succeeds, the found proof is displayed in the InfoView. It is a complete proof that is able to reduce the remaining goals to ``No goals''.}
  \vspace{-8mm}
  \label{fig:search_proof}
\end{figure*}

\paragraph{Proof Search with LLMs.}

To address this problem, \textsc{search\_proof} augments \textsc{aesop}'s rule set with goal-dependent tactics generated by \textsc{suggest\_tactics}. 
It allows the rule set to be customized for every goal, which makes \textsc{aesop} substantially more flexible. 
We make \textsc{search\_proof} a drop-in replacement of \textsc{aesop}: users can easily switch between \textsc{search\_proof} and the original white-box \textsc{aesop} by activating/deactivating an option for LLM-generated tactics.
The frontend is similar to \textsc{suggest\_tactics}, except that a complete proof is shown in the InfoView instead of individual tactics.
If \textsc{search\_proof} succeeds, the found proof must be correct, so the color of display carries no meaning.
Users can similarly click on the found proof to replace the \textsc{search\_proof} tactic.

\subsection{Selecting \& Annotating Premises}
\label{subsec:premise_selection}

Another challenging yet important task in theorem proving is to find relevant premises that can potentially advance a proof. 
Lean has a large mathematical library \citep{mathlib} in addition to the many premises already in Lean's source code and standard library. 
Searching for the right premises from all the libraries can be extremely hard and labor-intensive.
Some works thus try to ease this process by building dedicated library search engines for Lean\footnote{An example is Moogle: \url{https://www.moogle.ai/}. It supports searching for premises using English keywords.}.
Such engines can be helpful when users already know what mathematical premise they want to use, and just need to look up the formal equivalence in Lean.
Yet users may not always have a firm idea about what premises to use, and a library search can fail if the developer named a particular premise differently from what the users would.
The ability to select premises given proof goals is thus desired.

\paragraph{Premise Selection.}

A body of works tries to automate this process in Lean and other proof assistants \citep{alemi2017deepmath, mikuła2024magnushammer, wang2017premise}. 
The state-of-the-art tool for premise selection in Lean has been a random forest-based framework \citep{piotrowski2023machinelearned}. 
To bring more advanced neural architectures, we notice that premise selection is an extremely suitable task for a retrieval-augmented LLM such as ReProver, where a retrieval matrix (the \textit{premise embedding}) is trained to estimate the relevancy between a proof goal and candidate premises. 
Given a proof goal at inference time, we first encode the goal into a vector, and then perform a matrix-vector multiplication between the pre-computed premise embedding and the goal vector. 
The result is a vector of relevancy scores for the premises, from which we can select the highest ones. 
For matrix-vector multiplication, we need an efficient matrix multiplication library and a numpy matrix reader in C++. 
We adopt the matrix multiplication functions from CTranslate2 \citep{ct2}, and a fast numpy file reader in C++ from Libnpy \citep{libnpy}. 
We call these C++ functions from Lean again using FFI.

\paragraph{Premise Annotation.}

\begin{figure*}[t]
    \centering
    \vspace{-8mm}
    \includegraphics[width=1.0\linewidth]{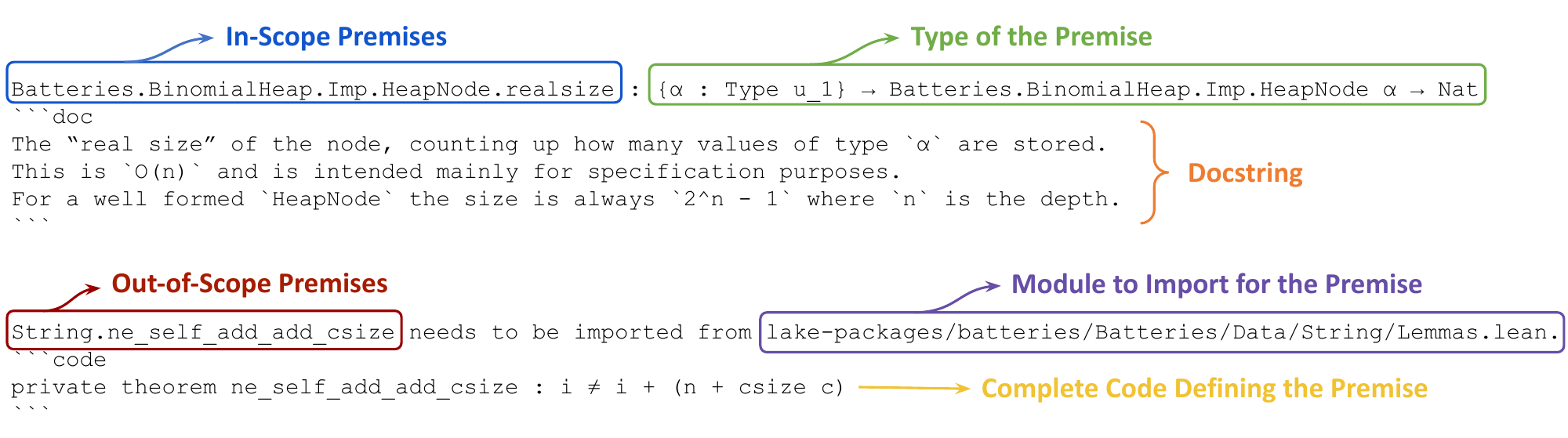}
    \vspace{-6mm}
    \caption{Two examples of annotated premises. The premises are selected based on proof goals, and then annotated according to whether they are in scope (i.e., all necessary modules imported, ready to be used) or not. The top one is in scope, annotated with its type information and docstring; the bottom one is out of scope, annotated with the module name that needs importing to use the premise, along with its complete definition code.}
    \vspace{-8mm}
    \label{fig:premises}
\end{figure*}

Premises are unlike tactics or proofs: we cannot autonomously verify whether a premise is useful, as there are many ways to apply the same premise in different tactics, which may yield different outcomes.
Thus, we help users more easily decide among selected premises by premise annotation via meta-programming.
We categorize each selected premise depending on whether its required module is already imported in the current environment.
If yes, the premise can readily be used.
We then show its type and docstring (if exists). 
Users can compare the premise type with current proof goals, and read useful explanations from the docstring about the mathematical meaning of a premise. 
Otherwise, if a premise cannot be directly used due to its module not imported, we provide users with the module name and the complete code that defines this premise.
The code helps users see if the definition of an out-of-scope premise is relevant and worth importing the additional module, and if so, knowing which module to import saves efforts.
The annotated premises are displayed in Lean's InfoView. 
Figure~\ref{fig:premises} shows 2 examples of annotated premises, one in scope and the other not.
The rest of the frontend looks the same as \textsc{suggest\_tactics} and \textsc{search\_proof}.

%% file: sections/experiments.tex
\section{Experiments}
\label{sec:experiments}

We empirically validate our hypothesis that human-AI collaboration is beneficial for theorem proving in Lean.
We focus on evaluating \textsc{suggest\_tactics} and \textsc{search\_proof}, as \textsc{select\_premises} is mainly a helping tool, and it remains a challenge in the field to effectively evaluate premise selection due to the inherent lack of ground truths.
We use \textsc{aesop} as a state-of-the-art rule-based baseline for proof search in Lean.
We compare \textsc{search\_proof} with \textsc{aesop} in two settings: (1) proving theorems fully autonomously (no human intervention) and (2) assisting humans in theorem proving. 
We also compare \textsc{search\_proof} with \textsc{suggest\_tactics}, to demonstrate the benefit of adding a proof search algorithm on top of tactic suggestion.

We investigate how effectively {\lc} can assist humans in theorem proving, in a similar paradigm as humans use copilots in software programming. 
That is, whenever a goal exists, humans first call the copilot to see if it can solve it directly. 
If not, humans attempt to proceed one step and try copilots again on the remaining goals. 
The above procedure is repeated until the copilots successfully solve the remaining goals at a certain step, or when humans solve all steps with no useful help from copilots. 
We investigate how much human effort can be automated by each proof automation tool, in such an iterative collaboration paradigm. 
The specific experiment design is as follows.

\paragraph{Dataset and Experimental Setup.}

We perform experiments on theorems from \textit{Mathematics in Lean} (MIL){\citep{avigad2020mathematics}}: a book for beginners to formalize and prove mathematical theorems in Lean. 
MIL is newly released on Github in May 2023, with parts of it made public for a short time in 2021.
Both potential sources do not lead to concerns about data contamination in our experiment, because their cutoff dates are earlier than ReProver's base model ByT5, and they were not part of ReProver's open-source finetuning data.
This may also explain why our tools usually come up with different proofs than ground truths in MIL in the experiment.

MIL contains theorem proving exercises that cover topics widely from sets and functions to topology, calculus, and measure theory. 
We evaluate on all 168 theorems in the book that have a tactic-style proof. 
Their proofs have 5.85 tactics on average (983 tactics in 168 theorems). 
The complete list of theorems, together with their detailed evaluation results, is presented in Appendix~\ref{sec:detailed}.

Each theorem comes with a ground-truth proof consisting of one or multiple tactics. 
To mimic a human user, we enter the ground truth tactics one by one. 
After each tactic, we try to prove the remaining goals using each of the automated tools: \textsc{search\_proof}, \textsc{aesop}, and \textsc{suggest\_tactics}. 
For \textsc{aesop}, we use it out of the box, without manually configuring the rule set. 
For \textsc{suggest\_tactics}, we say it proves a goal when one of the generated tactic suggestions can prove the goal. 
We record the number of tactics entered manually before the tool succeeds, so the number is zero if it can prove the original theorem fully autonomously without requiring any human-entered tactics. 

\paragraph{Results.}

\input{tables/results_overview}

Table~\ref{table:results_overview} summarizes experimental results.
\textsc{suggest\_tactics} generates the next \textit{one} tactic based on a proof goal.
\textsc{search\_proof} further combines \textit{single} tactic generation with a proof search algorithm and generates a complete proof.
\textsc{aesop} automates on the same level by searching for a complete proof, but its tactics are from a pre-defined list rather than generated by LLMs.
Our \textsc{search\_proof} can prove \textbf{63.7\%} (107 out of 168) theorems autonomously, which is significantly higher than \textsc{aesop} and \textsc{suggest\_tactics}.
When used to assist humans, \textsc{search\_proof} only requires an average of \textbf{2.08} manually-entered tactics, which compares favorably to \textsc{aesop} (3.86) and \textsc{suggest\_tactics} (3.11).
Finally, for each theorem, we calculate the percentage of proof steps that are automated by each of the three tools. 
Averaging the percentage over all tested theorems, we find that \textsc{search\_proof} can automate about \textbf{74.2\%} of the proof steps in a theorem, significantly more helpful than \textsc{suggest\_tactics} (58.4\%) and \textsc{aesop} (40.1\%). 
\textsc{Search\_proof} achieves a performance \textbf{27\%} better than \textsc{suggest\_tactics}, showing the effectiveness to add the proof search algorithm on top of our LLM-powered single tactic generation; and it's \textbf{85\%} better than the rule-based baseline \textsc{aesop}, showing the significantly beneficial role a LLM can play when assisting humans to prove theorems in Lean. 

%% file: tables/results_overview.tex
\begin{table*}[ht]
  \small
  \centering
  \resizebox{\textwidth}{!}{
  \begin{tabular}{@{}lccc@{}}
    \toprule
     Method & Avg. \# human-entered tactics ($\downarrow$) & \% theorems proved autonomously ($\uparrow$) & Avg. \% proof steps automated  ($\uparrow$) \\
    \midrule
    \textsc{aesop} & 3.86 & 24.4\% & 40.1\% \\
    \textsc{suggest\_tactics} & 3.10 & 45.2\% & 58.3\% \\
    \textsc{search\_proofs} & \textbf{2.08} & \textbf{63.7\%} & \textbf{74.2\%} \\
    \bottomrule
  \end{tabular}
  }
  \caption{Performance of \textsc{suggest\_tactics}, \textsc{aesop} and \textsc{search\_proof} on proving all theorems in ``Mathematics in Lean'' \citep{avigad2020mathematics} that have tactic-style proofs. \textsc{Search\_proof} outperforms both baselines in proving theorems autonomously and in assisting human users, requiring fewer tactics entered by humans.}
  \vspace{-12pt}
  \label{table:results_overview}
\end{table*}

%% file: sections/conclusion.tex
\section{Conclusion}
\label{sec:conclusion}

We have introduced {\lc}: a framework for running LLM inference natively in Lean. 
Using {\lc}, we have built LLM-based proof automation tools for generating tactic suggestions (\textsc{suggest\_tactics}), searching for proofs (\textsc{search\_proof}), and selecting premises (\textsc{select\_premises}). 
{\lc} also provides general interfaces between LLMs and Lean, allowing users to bring their own models and/or build other proof automation tools. 
Experimental results on the \textit{Mathematics in Lean} textbook demonstrate the effectiveness of our method compared to existing rule-based proof automation in Lean (\textsc{aesop}).
When assisting humans, {\lc} requires only \textbf{2.08} manually-entered proof steps on average (3.86 required by \textsc{aesop}); when automating the theorem proving process, {\lc} automates \textbf{74.2\%} proof steps on average, \textbf{85\%} better than \textsc{aesop} (40.1\%). 
These results confirm the benefit of integrating LLMs into the theorem proving pipeline in Lean.
We open source all code and artifacts to facilitate future research, and we hope to see more LLM-based proof automation built upon {\lc} to help create more high-quality formal data, which would in turn enhance LLMs' capability in formal math.

%% file: sections/artifacts.tex
\section{Code \& Artifacts}
\label{sec:artifacts}

We open source \textit{all code and artifacts} in our Github repository: \url{https://github.com/lean-dojo/LeanCopilot}, under a permissive MIT license.
We hope that the three automation tools we have built would be able to ease theorem proving in Lean, and we hope the general neuro-symbolic framework of {\lc} would encourage developers to continue building more automation tools that accelerate theorem proving in Lean.

We also release all evaluation code for our experiments presented in Section~\ref{sec:experiments} and Appendix~\ref{sec:detailed}, for full transparency.
The code is at \url{https://github.com/Peiyang-Song/mathematics_in_lean/tree/full-scale-experiment}.

%% file: sections/generalize.tex
\section{Generalizing {\lc} with Low-Level Interfaces}
\label{sec:generalize}

Section~\ref{sec:tools} has introduced the proof automation tools that we build for three important tasks in theorem proving in Lean -- tactic suggestion, proof search, and premise selection. 
In this section, we generalize the usability of our {\lc} framework for developers or advanced users who wish to engage in more low-level development and build their own proof automation tools. 
We provide two low-level interfaces in {\lc}, which correspond to two prevalent use cases of LLM inference: text-to-text generation and text-to-vector encoding.
We introduce the interfaces in this section of Appendix, as they are not immediate tools from {\lc} but could be very valuable for developers who wish to build other proof automation tools in Lean.

\paragraph{Text-to-Text Generation.}

\begin{figure*}[t]
  \centering
    \includegraphics[width=1.0\linewidth]{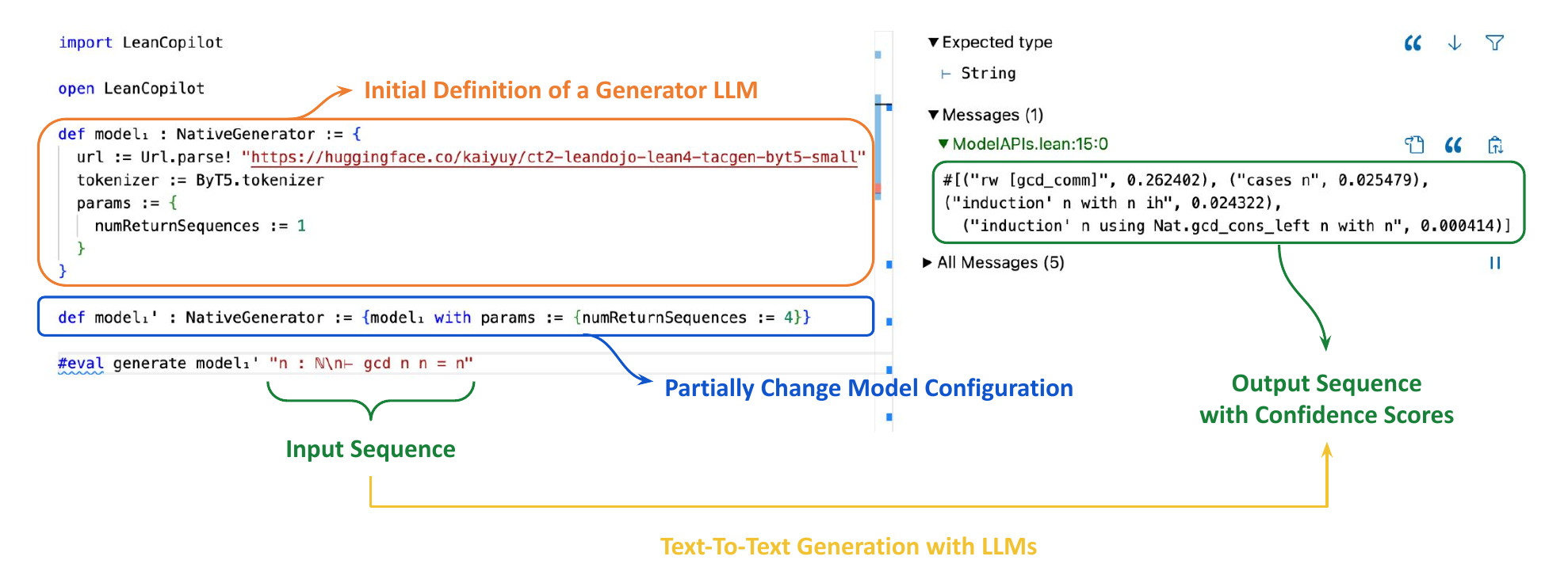}
  \caption{We provide an interface for text-to-text generation. Users can specify a model to use and configure it for generation. The model is not fixed after initial definition. If users would like to change part of its configuration, they can easily change certain the corresponding fields without the need to declare a new one. In this example, the user defines \texttt{$model_1'$} by changing the number of return sequences in \texttt{$model_1$} from 1 to 4. As a result, in the InfoView, $4$ sequences are returned for the \texttt{\#eval} statement when \texttt{$model_1'$} is used for generation.}
  \label{fig:text_to_text}
\end{figure*}

One major usage of LLM inference is text-to-text generation. 
Given a text sequence as input, LLMs generate an output text sequence. 
This is the mechanism behind \textsc{suggest\_tactics} and \textsc{search\_proof}, where our input text sequence is the remaining goals and our output sequence is a \textit{single} tactic suggestion. 
The reason why we can get multiple suggestions at once is due to beam search, which returns the top-$k$ (where $k$ is a hyperparameter controlling the number of output sequences to be returned) output sequences. 

With this interface, users can directly bring their own model and configure it for generation, as shown in the definition of \texttt{$model_1$} in Fig.~\ref{fig:text_to_text} below. 
A \texttt{url} should be provided to load a pretrained LLM from HuggingFace or other platforms, and a \texttt{tokenizer} can be specified to tokenize inputs to the LLM. 
Additional configurations used for generation can be added in the \texttt{params} field.

Naturally, when using a LLM, users sometimes would like to try different configurations for generation, which requires changing part of a configuration but not all of it. 
We support this need by allowing users to change part of an existing model configuration directly, with simple syntax shown in \texttt{$model_1'$} in Fig.~\ref{fig:text_to_text}. 
In this case, the user would like to use the same model as \texttt{$model_1$} but with the number of return sequences set to 4 rather than 1. 
The infoview (Figure~\ref{fig:text_to_text} right) shows the successfully generated text sequences. 
The output is from the \texttt{\#eval} statement in the codes (Figure~\ref{fig:text_to_text} left) which uses \texttt{$model_1'$} for generation. 
Changing any other field shares the same syntax.

\paragraph{Text-to-Vector Encoding.}

\begin{figure*}[t]
  \centering
    \includegraphics[width=1.0\linewidth]{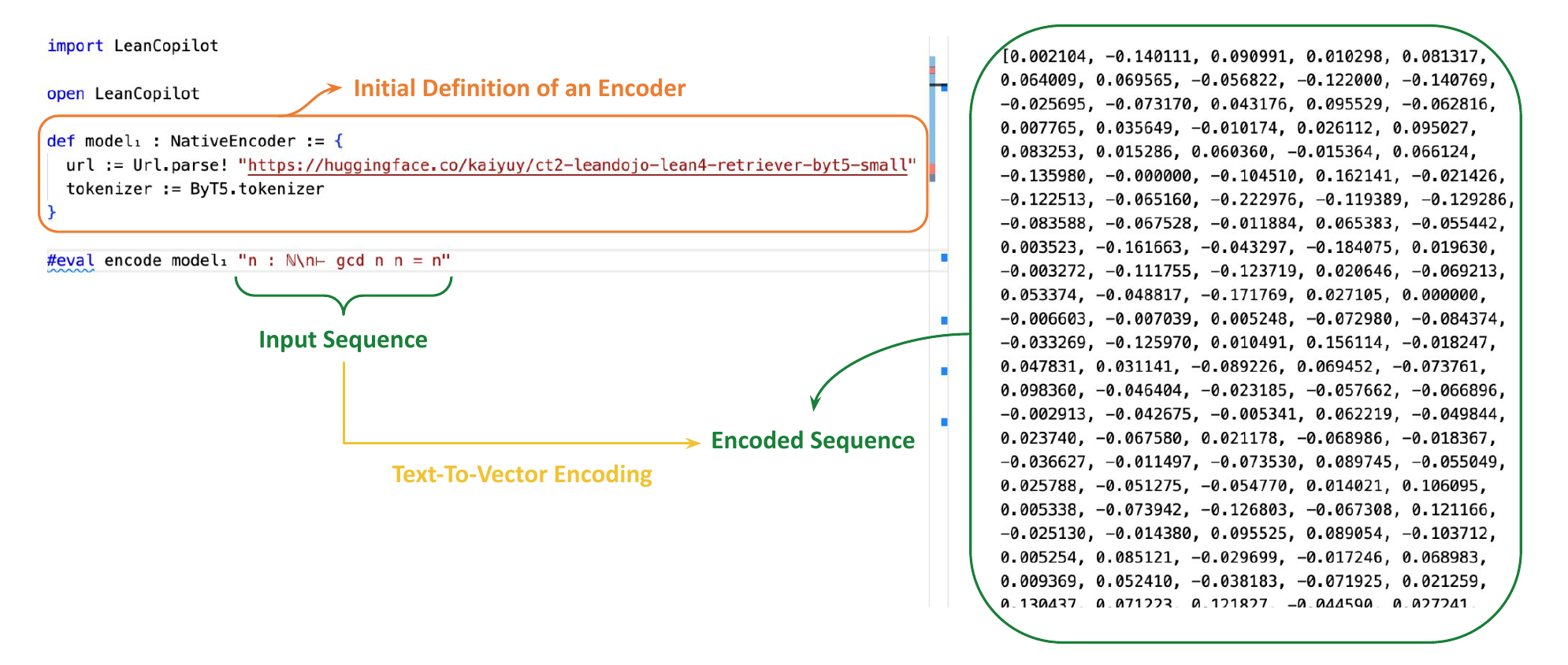}
  \caption{We provide a second interface for text-to-vector encoding. Users can bring their own model and configure it for encoding. A given string as input will be encoded into a vector of floats.}
  \label{fig:text_to_vector}
\end{figure*}

Another popular use of LLM inference is text-to-vector encoding, which encodes an input text sequence into a word vector embedding. 
This is part of the mechanism behind \textsc{select\_premises}, where our input is the remaining goals and our output is an encoded vector embedding for the input. 
We multiply this goal vector embedding with the pre-computed premise embedding from ReProver to get the scores of candidate premises.
With this interface, users can similarly bring their own model and provide any input sequence to be encoded, as shown in Fig.~\ref{fig:text_to_vector}.

Such two low-level interfaces together provide users with the freedom to bring their own models and configure the generation or encoding process to their need. 
Furthermore, while our native interfaces are theoretically able to perform text-to-text generation and text-to-vector encoding for any model, we provide the additional flexibility for users to use a server process instead.
This can be valuable especially when users are working on a project interfacing directly between Lean and another programming language like Python.

\paragraph{Running LLMs in Lean through Server Processes.}

To run LLMs in Lean through server processes, users can declare an \texttt{ExternalGenerator}, where they specify a model to use, a host name, and a port number. 
Then once they open the corresponding server in the command line, the specified model will be able to perform text-to-text generation, and display the outputs on the infoview just like the native generators do in Fig.~\ref{fig:text_to_text}. 
Text-to-vector encoding can be performed in the same way, by declaring an \texttt{ExternalEncoder}.
In principle, users can bring any models using {\lc} through \texttt{ExternalGenerator} and \texttt{ExternalEncoder}. 
In order to use them, users just need to expose certain APIs such as model name, input sequence, generation prefix, etc. 
We provide examples of using this Python API server in our codebase, which current include running models from OpenAI's GPT family \citep{openai2024gpt4technicalreport}, Anthropic's Claude family \citep{TheC3}, Google's Gemini family \citep{geminiteam2024geminifamilyhighlycapable}, InternLM2 \citep{cai2024internlm2technicalreport}, etc.

%% file: sections/detailed.tex
\section{Detailed Experimental Results}
\label{sec:detailed}

We show in Table~\ref{table:results_details} our complete experimental results on all theorems from Mathematics in Lean \citep{avigad2020mathematics} that have tactic-style proofs. 
The aggregated statistics are in Table~\ref{table:results_overview}.
We make all evaluation code public for full transparency, at \url{https://github.com/Peiyang-Song/mathematics_in_lean/tree/full-scale-experiment}.

For each theorem, the link in the table below directs to the file that contains ground-truth proofs for individual theorems.
The corresponding exercises are in the same folder of the repository, under the same name excluding the ``solution'' phrase.
Our experiment details are documented in the original ``problem'' files, while the ``solution'' files with ground-truth proofs are left intact.
In the ``problem'' files, we leave in comments \textsc{suggest\_tactics}, \textsc{aesop}, and \textsc{search\_proof}, at the position where they automated the rest of the proof.
That is, if one of the three automation tactics appears particularly in the next line after the proof, that means the automation tactic has not helped with the proof at all, and all steps are written by users.

\textsc{Aesop} is a white-box rule-based tool, so uncommenting the line \textsc{aesop} at the given position will guarantee to finish the rest of the proof, which means our evaluation at each theorem can be exactly reproduced.
\textsc{suggest\_tactics} and \textsc{search\_proof} uses neural networks under the hood, thus inherently probabilistic.
It is possible that when running those two tactics again, a different tactic/proof would be generated, possibly leading to slightly different evaluation results.
Therefore, to ensure transparency, in addition to the comments showing the exact positions where those two tools are able to automate the rest of the proof, we also leave in comments the tactic/proof that they generated during our run of this experiment.

\input{tables/results_details}

%% file: tables/results_details.tex
\begin{longtable}{@{}lccccccc@{}}
  
  \caption{Results on all theorems from Mathematics in Lean \citep{avigad2020mathematics} that have tactic-based proofs. The ``\# Tactics'' column shows the number of tactics in the ground truth proof. The ``\# Human tactics'' columns are the number of human-entered tactics required for the automated tool to finish the proof. The ``Auto'' columns show whether the tool can prove the theorem without humans, i.e., requiring zero human-entered tactics.}\\[1ex]
  
  \toprule
  
  \multirow{2}{*}{Theorem} & \multirow{2}{*}{\# Tactics} & \multicolumn{2}{c}{\textsc{aesop}} & \multicolumn{2}{c}{\textsc{suggest\_tactics}} & \multicolumn{2}{c}{\textsc{search\_proof}} \\
   & & \# Human tactics & Auto & \# Human tactics & Auto & \# Human tactics & Auto \\
  \midrule
  \endfirsthead

  \multicolumn{8}{c}%
  {\tablename\ \thetable\ -- \textit{Continued from previous page}} \\
  \toprule
  \multirow{2}{*}{Theorem} & \multirow{2}{*}{\# Tactics} & \multicolumn{2}{c}{\textsc{aesop}} & \multicolumn{2}{c}{\textsc{suggest\_tactics}} & \multicolumn{2}{c}{\textsc{search\_proof}} \\
   & & \# Human tactics & Auto & \# Human tactics & Auto & \# Human tactics & Auto \\
  \midrule
  \endhead
  
  \midrule \multicolumn{8}{r}{\textit{Continued on next page}} \\
  \endfoot
  
  \bottomrule
  \endlastfoot
  
    \href{https://github.com/leanprover-community/mathematics_in_lean/blob/53e574e0cccd781578aa4f43e2f8633fd17c09f9/MIL/C02_Basics/solutions/Solutions_S01_Calculating.lean#L3}{C02\_S01:3} & 3 & 3 & No & \textbf{0} & \textbf{Yes} & \textbf{0} & \textbf{Yes} \\
    \href{https://github.com/leanprover-community/mathematics_in_lean/blob/53e574e0cccd781578aa4f43e2f8633fd17c09f9/MIL/C02_Basics/solutions/Solutions_S01_Calculating.lean#L8}{C02\_S01:8} & 3 & 3 & No & \textbf{0} & \textbf{Yes} & \textbf{0} & \textbf{Yes} \\
    \href{https://github.com/leanprover-community/mathematics_in_lean/blob/53e574e0cccd781578aa4f43e2f8633fd17c09f9/MIL/C02_Basics/solutions/Solutions_S01_Calculating.lean#L13}{C02\_S01:13} & 2 & 2 & No & \textbf{0} & \textbf{Yes} & \textbf{0} & \textbf{Yes} \\
    \href{https://github.com/leanprover-community/mathematics_in_lean/blob/53e574e0cccd781578aa4f43e2f8633fd17c09f9/MIL/C02_Basics/solutions/Solutions_S01_Calculating.lean#L17}{C02\_S01:17} & 3 & 3 & No & \textbf{0} & \textbf{Yes} & \textbf{0} & \textbf{Yes} \\
    \href{https://github.com/leanprover-community/mathematics_in_lean/blob/53e574e0cccd781578aa4f43e2f8633fd17c09f9/MIL/C02_Basics/solutions/Solutions_S01_Calculating.lean#L22}{C02\_S01:22} & 3 & 3 & No & \textbf{0} & \textbf{Yes} & \textbf{0} & \textbf{Yes} \\
    \href{https://github.com/leanprover-community/mathematics_in_lean/blob/53e574e0cccd781578aa4f43e2f8633fd17c09f9/MIL/C02_Basics/solutions/Solutions_S01_Calculating.lean#L28}{C02\_S01:28} & 4 & 3 & No & \textbf{0} & \textbf{Yes} & \textbf{0} & \textbf{Yes} \\
    \href{https://github.com/leanprover-community/mathematics_in_lean/blob/53e574e0cccd781578aa4f43e2f8633fd17c09f9/MIL/C02_Basics/solutions/Solutions_S02_Proving_Identities_in_Algebraic_Structures.lean#L8}{C02\_S02:8} & 1 & \textbf{0} & \textbf{Yes} & \textbf{0} & \textbf{Yes} & \textbf{0} & \textbf{Yes} \\
    \href{https://github.com/leanprover-community/mathematics_in_lean/blob/53e574e0cccd781578aa4f43e2f8633fd17c09f9/MIL/C02_Basics/solutions/Solutions_S02_Proving_Identities_in_Algebraic_Structures.lean#L11}{C02\_S02:11} & 1 & \textbf{0} & \textbf{Yes} & \textbf{0} & \textbf{Yes} & \textbf{0} & \textbf{Yes} \\
    \href{https://github.com/leanprover-community/mathematics_in_lean/blob/53e574e0cccd781578aa4f43e2f8633fd17c09f9/MIL/C02_Basics/solutions/Solutions_S02_Proving_Identities_in_Algebraic_Structures.lean#L14}{C02\_S02:14} & 1 & \textbf{0} & \textbf{Yes} & \textbf{0} & \textbf{Yes} & \textbf{0} & \textbf{Yes} \\
    \href{https://github.com/leanprover-community/mathematics_in_lean/blob/53e574e0cccd781578aa4f43e2f8633fd17c09f9/MIL/C02_Basics/solutions/Solutions_S02_Proving_Identities_in_Algebraic_Structures.lean#L17}{C02\_S02:17} & 2 & \textbf{0} & \textbf{Yes} & \textbf{0} & \textbf{Yes} & \textbf{0} & \textbf{Yes} \\
    \href{https://github.com/leanprover-community/mathematics_in_lean/blob/53e574e0cccd781578aa4f43e2f8633fd17c09f9/MIL/C02_Basics/solutions/Solutions_S02_Proving_Identities_in_Algebraic_Structures.lean#L21}{C02\_S02:21} & 1 & 1 & No & \textbf{0} & \textbf{Yes} & \textbf{0} & \textbf{Yes} \\
    \href{https://github.com/leanprover-community/mathematics_in_lean/blob/53e574e0cccd781578aa4f43e2f8633fd17c09f9/MIL/C02_Basics/solutions/Solutions_S02_Proving_Identities_in_Algebraic_Structures.lean#L24}{C02\_S02:24} & 3 & 3 & No & \textbf{0} & \textbf{Yes} & \textbf{0} & \textbf{Yes} \\
    \href{https://github.com/leanprover-community/mathematics_in_lean/blob/53e574e0cccd781578aa4f43e2f8633fd17c09f9/MIL/C02_Basics/solutions/Solutions_S02_Proving_Identities_in_Algebraic_Structures.lean#L29}{C02\_S02:29} & 2 & \textbf{0} & \textbf{Yes} & \textbf{0} & \textbf{Yes} & \textbf{0} & \textbf{Yes} \\
    \href{https://github.com/leanprover-community/mathematics_in_lean/blob/53e574e0cccd781578aa4f43e2f8633fd17c09f9/MIL/C02_Basics/solutions/Solutions_S02_Proving_Identities_in_Algebraic_Structures.lean#L33}{C02\_S02:33} & 2 & \textbf{0} & \textbf{Yes} & \textbf{0} & \textbf{Yes} & \textbf{0} & \textbf{Yes} \\
    \href{https://github.com/leanprover-community/mathematics_in_lean/blob/53e574e0cccd781578aa4f43e2f8633fd17c09f9/MIL/C02_Basics/solutions/Solutions_S02_Proving_Identities_in_Algebraic_Structures.lean#L42}{C02\_S02:42} & 1 & \textbf{0} & \textbf{Yes} & \textbf{0} & \textbf{Yes} & \textbf{0} & \textbf{Yes} \\
    \href{https://github.com/leanprover-community/mathematics_in_lean/blob/53e574e0cccd781578aa4f43e2f8633fd17c09f9/MIL/C02_Basics/solutions/Solutions_S02_Proving_Identities_in_Algebraic_Structures.lean#L45}{C02\_S02:45} & 1 & 1 & No & \textbf{0} & \textbf{Yes} & \textbf{0} & \textbf{Yes} \\
    \href{https://github.com/leanprover-community/mathematics_in_lean/blob/53e574e0cccd781578aa4f43e2f8633fd17c09f9/MIL/C02_Basics/solutions/Solutions_S02_Proving_Identities_in_Algebraic_Structures.lean#L48}{C02\_S02:48} & 1 & 1 & No & \textbf{0} & \textbf{Yes} & \textbf{0} & \textbf{Yes} \\
    \href{https://github.com/leanprover-community/mathematics_in_lean/blob/53e574e0cccd781578aa4f43e2f8633fd17c09f9/MIL/C02_Basics/solutions/Solutions_S02_Proving_Identities_in_Algebraic_Structures.lean#L58}{C02\_S02:58} & 2 & \textbf{0} & \textbf{Yes} & \textbf{0} & \textbf{Yes} & \textbf{0} & \textbf{Yes} \\
    \href{https://github.com/leanprover-community/mathematics_in_lean/blob/53e574e0cccd781578aa4f43e2f8633fd17c09f9/MIL/C02_Basics/solutions/Solutions_S02_Proving_Identities_in_Algebraic_Structures.lean#L63}{C02\_S02:63} & 1 & \textbf{0} & \textbf{Yes} & \textbf{0} & \textbf{Yes} & \textbf{0} & \textbf{Yes} \\
    \href{https://github.com/leanprover-community/mathematics_in_lean/blob/53e574e0cccd781578aa4f43e2f8633fd17c09f9/MIL/C02_Basics/solutions/Solutions_S02_Proving_Identities_in_Algebraic_Structures.lean#L63}{C02\_S02:66} & 1 & \textbf{0} & \textbf{Yes} & \textbf{0} & \textbf{Yes} & \textbf{0} & \textbf{Yes} \\
    \href{https://github.com/leanprover-community/mathematics_in_lean/blob/53e574e0cccd781578aa4f43e2f8633fd17c09f9/MIL/C02_Basics/solutions/Solutions_S03_Using_Theorems_and_Lemmas.lean#L7}{C02\_S03:7} & 3 & 3 & No & \textbf{0} & \textbf{Yes} & \textbf{0} & \textbf{Yes} \\
    \href{https://github.com/leanprover-community/mathematics_in_lean/blob/53e574e0cccd781578aa4f43e2f8633fd17c09f9/MIL/C02_Basics/solutions/Solutions_S03_Using_Theorems_and_Lemmas.lean#L12}{C02\_S03:12} & 3 & \textbf{0} & \textbf{Yes} & \textbf{0} & \textbf{Yes} & \textbf{0} & \textbf{Yes} \\
    \href{https://github.com/leanprover-community/mathematics_in_lean/blob/53e574e0cccd781578aa4f43e2f8633fd17c09f9/MIL/C02_Basics/solutions/Solutions_S03_Using_Theorems_and_Lemmas.lean#L24}{C02\_S03:24} & 3 & 2 & No & \textbf{0} & \textbf{Yes} & \textbf{0} & \textbf{Yes} \\
    \href{https://github.com/leanprover-community/mathematics_in_lean/blob/53e574e0cccd781578aa4f43e2f8633fd17c09f9/MIL/C02_Basics/solutions/Solutions_S03_Using_Theorems_and_Lemmas.lean#L30}{C02\_S03:30} & 2 & 1 & No & \textbf{0} & \textbf{Yes} & \textbf{0} & \textbf{Yes} \\
    \href{https://github.com/leanprover-community/mathematics_in_lean/blob/53e574e0cccd781578aa4f43e2f8633fd17c09f9/MIL/C02_Basics/solutions/Solutions_S04_More_on_Order_and_Divisibility.lean#L13}{C02\_S04:13} & 5 & 5 & No & \textbf{0} & \textbf{Yes} & \textbf{0} & \textbf{Yes} \\
    \href{https://github.com/leanprover-community/mathematics_in_lean/blob/53e574e0cccd781578aa4f43e2f8633fd17c09f9/MIL/C02_Basics/solutions/Solutions_S04_More_on_Order_and_Divisibility.lean#L20}{C02\_S04:20} & 19 & 8 & No & 2 & No & 8 & No \\
    \href{https://github.com/leanprover-community/mathematics_in_lean/blob/53e574e0cccd781578aa4f43e2f8633fd17c09f9/MIL/C02_Basics/solutions/Solutions_S04_More_on_Order_and_Divisibility.lean#L41}{C02\_S04:41} & 5 & \textbf{0} & \textbf{Yes} & \textbf{0} & \textbf{Yes} & \textbf{0} & \textbf{Yes} \\
    \href{https://github.com/leanprover-community/mathematics_in_lean/blob/53e574e0cccd781578aa4f43e2f8633fd17c09f9/MIL/C02_Basics/solutions/Solutions_S04_More_on_Order_and_Divisibility.lean#L48}{C02\_S04:48} & 9 & 5 & No & \textbf{0} & \textbf{Yes} & \textbf{0} & \textbf{Yes} \\
    \href{https://github.com/leanprover-community/mathematics_in_lean/blob/53e574e0cccd781578aa4f43e2f8633fd17c09f9/MIL/C02_Basics/solutions/Solutions_S04_More_on_Order_and_Divisibility.lean#L89}{C02\_S04:89} & 8 & 7 & No & 6 & No & 4 & No \\
    \href{https://github.com/leanprover-community/mathematics_in_lean/blob/53e574e0cccd781578aa4f43e2f8633fd17c09f9/MIL/C02_Basics/solutions/Solutions_S04_More_on_Order_and_Divisibility.lean#L109}{C02\_S04:109} & 5 & 5 & No & \textbf{0} & \textbf{Yes} & \textbf{0} & \textbf{Yes} \\
    \href{https://github.com/leanprover-community/mathematics_in_lean/blob/53e574e0cccd781578aa4f43e2f8633fd17c09f9/MIL/C02_Basics/solutions/Solutions_S05_Proving_Facts_about_Algebraic_Structures.lean#L8}{C02\_S05:8} & 5 & 5 & No & \textbf{0} & \textbf{Yes} & \textbf{0} & \textbf{Yes} \\
    \href{https://github.com/leanprover-community/mathematics_in_lean/blob/53e574e0cccd781578aa4f43e2f8633fd17c09f9/MIL/C02_Basics/solutions/Solutions_S05_Proving_Facts_about_Algebraic_Structures.lean#L15}{C02\_S05:15} & 19 & 18 & No & \textbf{0} & \textbf{Yes} & \textbf{0} & \textbf{Yes} \\
    \href{https://github.com/leanprover-community/mathematics_in_lean/blob/53e574e0cccd781578aa4f43e2f8633fd17c09f9/MIL/C02_Basics/solutions/Solutions_S05_Proving_Facts_about_Algebraic_Structures.lean#L36}{C02\_S05:36} & 5 & 5 & No & \textbf{0} & \textbf{Yes} & \textbf{0} & \textbf{Yes} \\
    \href{https://github.com/leanprover-community/mathematics_in_lean/blob/53e574e0cccd781578aa4f43e2f8633fd17c09f9/MIL/C02_Basics/solutions/Solutions_S05_Proving_Facts_about_Algebraic_Structures.lean#L43}{C02\_S05:43} & 19 & 17 & No & \textbf{0} & \textbf{Yes} & \textbf{0} & \textbf{Yes} \\
    \href{https://github.com/leanprover-community/mathematics_in_lean/blob/53e574e0cccd781578aa4f43e2f8633fd17c09f9/MIL/C02_Basics/solutions/Solutions_S05_Proving_Facts_about_Algebraic_Structures.lean#L64}{C02\_S05:64} & 5 & \textbf{0} & \textbf{Yes} & \textbf{0} & \textbf{Yes} & \textbf{0} & \textbf{Yes} \\
    \href{https://github.com/leanprover-community/mathematics_in_lean/blob/53e574e0cccd781578aa4f43e2f8633fd17c09f9/MIL/C02_Basics/solutions/Solutions_S05_Proving_Facts_about_Algebraic_Structures.lean#L71}{C02\_S05:71} & 5 & \textbf{0} & \textbf{Yes} & \textbf{0} & \textbf{Yes} & \textbf{0} & \textbf{Yes} \\
    \href{https://github.com/leanprover-community/mathematics_in_lean/blob/53e574e0cccd781578aa4f43e2f8633fd17c09f9/MIL/C02_Basics/solutions/Solutions_S05_Proving_Facts_about_Algebraic_Structures.lean#L108}{C02\_S05:108} & 2 & \textbf{0} & \textbf{Yes} & \textbf{0} & \textbf{Yes} & \textbf{0} & \textbf{Yes} \\
    \href{https://github.com/leanprover-community/mathematics_in_lean/blob/53e574e0cccd781578aa4f43e2f8633fd17c09f9/MIL/C02_Basics/solutions/Solutions_S05_Proving_Facts_about_Algebraic_Structures.lean#L112}{C02\_S05:112} & 2 & \textbf{0} & \textbf{Yes} & \textbf{0} & \textbf{Yes} & \textbf{0} & \textbf{Yes} \\
    \href{https://github.com/leanprover-community/mathematics_in_lean/blob/53e574e0cccd781578aa4f43e2f8633fd17c09f9/MIL/C02_Basics/solutions/Solutions_S05_Proving_Facts_about_Algebraic_Structures.lean#L127}{C02\_S05:127} & 4 & 4 & No & \textbf{0} & \textbf{Yes} & \textbf{0} & \textbf{Yes} \\
    \href{https://github.com/leanprover-community/mathematics_in_lean/blob/53e574e0cccd781578aa4f43e2f8633fd17c09f9/MIL/C03_Logic/solutions/Solutions_S01_Implication_and_the_Universal_Quantifier.lean#L6}{C03\_S01:6} & 2 & 2 & No & 1 & No & \textbf{0} & \textbf{Yes} \\
    \href{https://github.com/leanprover-community/mathematics_in_lean/blob/53e574e0cccd781578aa4f43e2f8633fd17c09f9/MIL/C03_Logic/solutions/Solutions_S01_Implication_and_the_Universal_Quantifier.lean#L30}{C03\_S01:30} & 6 & 5 & No & 1 & No & 1 & No \\
    \href{https://github.com/leanprover-community/mathematics_in_lean/blob/53e574e0cccd781578aa4f43e2f8633fd17c09f9/MIL/C03_Logic/solutions/Solutions_S01_Implication_and_the_Universal_Quantifier.lean#L50}{C03\_S01:50} & 3 & 2 & No & \textbf{0} & \textbf{Yes} & \textbf{0} & \textbf{Yes} \\
    \href{https://github.com/leanprover-community/mathematics_in_lean/blob/53e574e0cccd781578aa4f43e2f8633fd17c09f9/MIL/C03_Logic/solutions/Solutions_S01_Implication_and_the_Universal_Quantifier.lean#L58}{C03\_S01:58} & 4 & 1 & No & \textbf{0} & \textbf{Yes} & \textbf{0} & \textbf{Yes} \\
    \href{https://github.com/leanprover-community/mathematics_in_lean/blob/53e574e0cccd781578aa4f43e2f8633fd17c09f9/MIL/C03_Logic/solutions/Solutions_S01_Implication_and_the_Universal_Quantifier.lean#L73}{C03\_S01:73} & 2 & 2 & No & 2 & No & 2 & No \\
    \href{https://github.com/leanprover-community/mathematics_in_lean/blob/53e574e0cccd781578aa4f43e2f8633fd17c09f9/MIL/C03_Logic/solutions/Solutions_S01_Implication_and_the_Universal_Quantifier.lean#L80}{C03\_S01:80} & 3 & 3 & No & 3 & No & 3 & No \\
    \href{https://github.com/leanprover-community/mathematics_in_lean/blob/53e574e0cccd781578aa4f43e2f8633fd17c09f9/MIL/C03_Logic/solutions/Solutions_S01_Implication_and_the_Universal_Quantifier.lean#L85}{C03\_S01:85} & 3 & 3 & No & 3 & No & 3 & No \\
    \href{https://github.com/leanprover-community/mathematics_in_lean/blob/53e574e0cccd781578aa4f43e2f8633fd17c09f9/MIL/C03_Logic/solutions/Solutions_S01_Implication_and_the_Universal_Quantifier.lean#L96}{C03\_S01:96} & 4 & 1 & No & \textbf{0} & \textbf{Yes} & \textbf{0} & \textbf{Yes} \\
    \href{https://github.com/leanprover-community/mathematics_in_lean/blob/53e574e0cccd781578aa4f43e2f8633fd17c09f9/MIL/C03_Logic/solutions/Solutions_S01_Implication_and_the_Universal_Quantifier.lean#L127}{C03\_S01:127} & 2 & 1 & No & \textbf{0} & \textbf{Yes} & \textbf{0} & \textbf{Yes} \\
    \href{https://github.com/leanprover-community/mathematics_in_lean/blob/53e574e0cccd781578aa4f43e2f8633fd17c09f9/MIL/C03_Logic/solutions/Solutions_S01_Implication_and_the_Universal_Quantifier.lean#L134}{C03\_S01:134} & 4 & 1 & No & \textbf{0} & \textbf{Yes} & \textbf{0} & \textbf{Yes} \\
    \href{https://github.com/leanprover-community/mathematics_in_lean/blob/53e574e0cccd781578aa4f43e2f8633fd17c09f9/MIL/C03_Logic/solutions/Solutions_S02_The_Existential_Quantifier.lean#L28}{C03\_S02:28} & 5 & 5 & No & 4 & No & 4 & No \\
    \href{https://github.com/leanprover-community/mathematics_in_lean/blob/53e574e0cccd781578aa4f43e2f8633fd17c09f9/MIL/C03_Logic/solutions/Solutions_S02_The_Existential_Quantifier.lean#L35}{C03\_S02:35} & 4 & 4 & No & 3 & No & 3 & No \\
    \href{https://github.com/leanprover-community/mathematics_in_lean/blob/53e574e0cccd781578aa4f43e2f8633fd17c09f9/MIL/C03_Logic/solutions/Solutions_S02_The_Existential_Quantifier.lean#L51}{C03\_S02:51} & 3 & 3 & No & \textbf{0} & \textbf{Yes} & \textbf{0} & \textbf{Yes} \\
    \href{https://github.com/leanprover-community/mathematics_in_lean/blob/53e574e0cccd781578aa4f43e2f8633fd17c09f9/MIL/C03_Logic/solutions/Solutions_S02_The_Existential_Quantifier.lean#L62}{C03\_S02:62} & 3 & 3 & No & 2 & No & 2 & No \\
    \href{https://github.com/leanprover-community/mathematics_in_lean/blob/53e574e0cccd781578aa4f43e2f8633fd17c09f9/MIL/C03_Logic/solutions/Solutions_S02_The_Existential_Quantifier.lean#L79}{C03\_S02:79} & 4 & 3 & No & \textbf{0} & \textbf{Yes} & \textbf{0} & \textbf{Yes} \\
    \href{https://github.com/leanprover-community/mathematics_in_lean/blob/53e574e0cccd781578aa4f43e2f8633fd17c09f9/MIL/C03_Logic/solutions/Solutions_S03_Negation.lean#L23}{C03\_S03:23} & 4 & 4 & No & 3 & No & 3 & No \\
    \href{https://github.com/leanprover-community/mathematics_in_lean/blob/53e574e0cccd781578aa4f43e2f8633fd17c09f9/MIL/C03_Logic/solutions/Solutions_S03_Negation.lean#L29}{C03\_S03:29} & 3 & 3 & No & 2 & No & 2 & No \\
    \href{https://github.com/leanprover-community/mathematics_in_lean/blob/53e574e0cccd781578aa4f43e2f8633fd17c09f9/MIL/C03_Logic/solutions/Solutions_S03_Negation.lean#L34}{C03\_S03:34} & 4 & 4 & No & \textbf{0} & \textbf{Yes} & \textbf{0} & \textbf{Yes} \\
    \href{https://github.com/leanprover-community/mathematics_in_lean/blob/53e574e0cccd781578aa4f43e2f8633fd17c09f9/MIL/C03_Logic/solutions/Solutions_S03_Negation.lean#L40}{C03\_S03:40} & 4 & 3 & No & 1 & No & \textbf{0} & \textbf{Yes} \\
    \href{https://github.com/leanprover-community/mathematics_in_lean/blob/53e574e0cccd781578aa4f43e2f8633fd17c09f9/MIL/C03_Logic/solutions/Solutions_S03_Negation.lean#L46}{C03\_S03:46} & 8 & 8 & No & 7 & No & 7 & No \\
    \href{https://github.com/leanprover-community/mathematics_in_lean/blob/53e574e0cccd781578aa4f43e2f8633fd17c09f9/MIL/C03_Logic/solutions/Solutions_S03_Negation.lean#L56}{C03\_S03:56} & 3 & 3 & No & 3 & No & \textbf{0} & \textbf{Yes} \\
    \href{https://github.com/leanprover-community/mathematics_in_lean/blob/53e574e0cccd781578aa4f43e2f8633fd17c09f9/MIL/C03_Logic/solutions/Solutions_S03_Negation.lean#L66}{C03\_S03:66} & 3 & \textbf{0} & \textbf{Yes} & \textbf{0} & \textbf{Yes} & \textbf{0} & \textbf{Yes} \\
    \href{https://github.com/leanprover-community/mathematics_in_lean/blob/53e574e0cccd781578aa4f43e2f8633fd17c09f9/MIL/C03_Logic/solutions/Solutions_S03_Negation.lean#L71}{C03\_S03:71} & 2 & \textbf{0} & \textbf{Yes} & \textbf{0} & \textbf{Yes} & \textbf{0} & \textbf{Yes} \\
    \href{https://github.com/leanprover-community/mathematics_in_lean/blob/53e574e0cccd781578aa4f43e2f8633fd17c09f9/MIL/C03_Logic/solutions/Solutions_S03_Negation.lean.lean#L75}{C03\_S03:75} & 4 & \textbf{0} & \textbf{Yes} & \textbf{0} & \textbf{Yes} & \textbf{0} & \textbf{Yes} \\
    \href{https://github.com/leanprover-community/mathematics_in_lean/blob/53e574e0cccd781578aa4f43e2f8633fd17c09f9/MIL/C03_Logic/solutions/Solutions_S03_Negation.lean#L81}{C03\_S03:81} & 2 & \textbf{0} & \textbf{Yes} & \textbf{0} & \textbf{Yes} & \textbf{0} & \textbf{Yes} \\
    \href{https://github.com/leanprover-community/mathematics_in_lean/blob/53e574e0cccd781578aa4f43e2f8633fd17c09f9/MIL/C03_Logic/solutions/Solutions_S03_Negation.lean#L85}{C03\_S03:85} & 2 & \textbf{0} & \textbf{Yes} & \textbf{0} & \textbf{Yes} & \textbf{0} & \textbf{Yes} \\
    \href{https://github.com/leanprover-community/mathematics_in_lean/blob/53e574e0cccd781578aa4f43e2f8633fd17c09f9/MIL/C03_Logic/solutions/Solutions_S03_Negation.lean#L94}{C03\_S03:94} & 10 & 4 & No & 4 & No & 4 & No \\
    \href{https://github.com/leanprover-community/mathematics_in_lean/blob/53e574e0cccd781578aa4f43e2f8633fd17c09f9/MIL/C03_Logic/solutions/Solutions_S03_Negation.lean#L105}{C03\_S03:105} & 3 & 1 & No & 1 & No & \textbf{0} & \textbf{Yes} \\
    \href{https://github.com/leanprover-community/mathematics_in_lean/blob/53e574e0cccd781578aa4f43e2f8633fd17c09f9/MIL/C03_Logic/solutions/Solutions_S04_Conjunction_and_Iff.lean#L7}{C03\_S04:7} & 6 & 6 & No & 6 & No & 6 & No \\
    \href{https://github.com/leanprover-community/mathematics_in_lean/blob/53e574e0cccd781578aa4f43e2f8633fd17c09f9/MIL/C03_Logic/solutions/Solutions_S04_Conjunction_and_Iff.lean#L15}{C03\_S04:15} & 13 & 13 & No & 12 & No & \textbf{0} & \textbf{Yes} \\
    \href{https://github.com/leanprover-community/mathematics_in_lean/blob/53e574e0cccd781578aa4f43e2f8633fd17c09f9/MIL/C03_Logic/solutions/Solutions_S04_Conjunction_and_Iff.lean#L34}{C03\_S04:34} & 9 & 7 & No & 8 & No & 7 & No \\
    \href{https://github.com/leanprover-community/mathematics_in_lean/blob/53e574e0cccd781578aa4f43e2f8633fd17c09f9/MIL/C03_Logic/solutions/Solutions_S04_Conjunction_and_Iff.lean#L49}{C03\_S04:49} & 3 & 2 & No & 2 & No & 2 & No \\
    \href{https://github.com/leanprover-community/mathematics_in_lean/blob/53e574e0cccd781578aa4f43e2f8633fd17c09f9/MIL/C03_Logic/solutions/Solutions_S04_Conjunction_and_Iff.lean#L58}{C03\_S04:58} & 14 & 14 & No & \textbf{0} & \textbf{Yes} & \textbf{0} & \textbf{Yes} \\
    \href{https://github.com/leanprover-community/mathematics_in_lean/blob/53e574e0cccd781578aa4f43e2f8633fd17c09f9/MIL/C03_Logic/solutions/Solutions_S04_Conjunction_and_Iff.lean#L80}{C03\_S04:80} & 3 & \textbf{0} & \textbf{Yes} & \textbf{0} & \textbf{Yes} & \textbf{0} & \textbf{Yes} \\
    \href{https://github.com/leanprover-community/mathematics_in_lean/blob/53e574e0cccd781578aa4f43e2f8633fd17c09f9/MIL/C03_Logic/solutions/Solutions_S04_Conjunction_and_Iff.lean#L85}{C03\_S04:85} & 7 & 7 & No & 7 & No & \textbf{0} & \textbf{Yes} \\
    \href{https://github.com/leanprover-community/mathematics_in_lean/blob/53e574e0cccd781578aa4f43e2f8633fd17c09f9/MIL/C03_Logic/solutions/Solutions_S04_Conjunction_and_Iff.lean#L60}{C03\_S04:60} & 4 & 4 & No & 3 & No & \textbf{0} & \textbf{Yes} \\
    \href{https://github.com/leanprover-community/mathematics_in_lean/blob/53e574e0cccd781578aa4f43e2f8633fd17c09f9/MIL/C03_Logic/solutions/Solutions_S05_Disjunction.lean#L12}{C03\_S05:12} & 4 & 4 & No & 3 & No & \textbf{0} & \textbf{Yes} \\
    \href{https://github.com/leanprover-community/mathematics_in_lean/blob/53e574e0cccd781578aa4f43e2f8633fd17c09f9/MIL/C03_Logic/solutions/Solutions_S05_Disjunction.lean#L18}{C03\_S05:18} & 4 & 4 & No & \textbf{0} & \textbf{Yes} & \textbf{0} & \textbf{Yes} \\
    \href{https://github.com/leanprover-community/mathematics_in_lean/blob/53e574e0cccd781578aa4f43e2f8633fd17c09f9/MIL/C03_Logic/solutions/Solutions_S05_Disjunction.lean#L24}{C03\_S05:24} & 5 & 5 & No & 5 & No & \textbf{0} & \textbf{Yes} \\
    \href{https://github.com/leanprover-community/mathematics_in_lean/blob/53e574e0cccd781578aa4f43e2f8633fd17c09f9/MIL/C03_Logic/solutions/Solutions_S05_Disjunction.lean#L31}{C03\_S05:31} & 20 & 16 & No & 16 & No & 10 & No \\
    \href{https://github.com/leanprover-community/mathematics_in_lean/blob/53e574e0cccd781578aa4f43e2f8633fd17c09f9/MIL/C03_Logic/solutions/Solutions_S05_Disjunction.lean#L52}{C03\_S05:52} & 18 & 15 & No & 10 & No & 1 & No \\
    \href{https://github.com/leanprover-community/mathematics_in_lean/blob/53e574e0cccd781578aa4f43e2f8633fd17c09f9/MIL/C03_Logic/solutions/Solutions_S05_Disjunction.lean#L76}{C03\_S05:76} & 1 & 1 & No & 1 & No & \textbf{0} & \textbf{Yes} \\
    \href{https://github.com/leanprover-community/mathematics_in_lean/blob/53e574e0cccd781578aa4f43e2f8633fd17c09f9/MIL/C03_Logic/solutions/Solutions_S05_Disjunction.lean#L79}{C03\_S05:79} & 9 & \textbf{0} & \textbf{Yes} & \textbf{0} & \textbf{Yes} & \textbf{0} & \textbf{Yes} \\
    \href{https://github.com/leanprover-community/mathematics_in_lean/blob/53e574e0cccd781578aa4f43e2f8633fd17c09f9/MIL/C03_Logic/solutions/Solutions_S05_Disjunction.lean#L90}{C03\_S05:90} & 9 & 9 & No & 7 & No & 4 & No \\
    \href{https://github.com/leanprover-community/mathematics_in_lean/blob/53e574e0cccd781578aa4f43e2f8633fd17c09f9/MIL/C03_Logic/solutions/Solutions_S05_Disjunction.lean#L105}{C03\_S05:105} & 9 & \textbf{0} & \textbf{Yes} & \textbf{0} & \textbf{Yes} & \textbf{0} & \textbf{Yes} \\
    \href{https://github.com/leanprover-community/mathematics_in_lean/blob/53e574e0cccd781578aa4f43e2f8633fd17c09f9/MIL/C03_Logic/solutions/Solutions_S05_Disjunction.lean#L116}{C03\_S05:116} & 9 & 9 & No & 7 & No & 5 & No \\
    \href{https://github.com/leanprover-community/mathematics_in_lean/blob/53e574e0cccd781578aa4f43e2f8633fd17c09f9/MIL/C03_Logic/solutions/Solutions_S05_Disjunction.lean#L129}{C03\_S05:129} & 12 & 3 & No & \textbf{0} & \textbf{Yes} & \textbf{0} & \textbf{Yes} \\
    \href{https://github.com/leanprover-community/mathematics_in_lean/blob/53e574e0cccd781578aa4f43e2f8633fd17c09f9/MIL/C03_Logic/solutions/Solutions_S06_Sequences_and_Convergence.lean#L16}{C03\_S06:16} & 10 & 10 & No & 10 & No & 10 & No \\
    \href{https://github.com/leanprover-community/mathematics_in_lean/blob/53e574e0cccd781578aa4f43e2f8633fd17c09f9/MIL/C03_Logic/solutions/Solutions_S06_Sequences_and_Convergence.lean#L36}{C03\_S06:36} & 14 & 14 & No & 14 & No & 14 & No \\
    \href{https://github.com/leanprover-community/mathematics_in_lean/blob/53e574e0cccd781578aa4f43e2f8633fd17c09f9/MIL/C03_Logic/solutions/Solutions_S06_Sequences_and_Convergence.lean#L56}{C03\_S06:56} & 4 & 4 & No & 4 & No & 4 & No \\
    \href{https://github.com/leanprover-community/mathematics_in_lean/blob/53e574e0cccd781578aa4f43e2f8633fd17c09f9/MIL/C03_Logic/solutions/Solutions_S06_Sequences_and_Convergence.lean#L68}{C03\_S06:68} & 11 & 11 & No & 11 & No & 11 & No \\
    \href{https://github.com/leanprover-community/mathematics_in_lean/blob/53e574e0cccd781578aa4f43e2f8633fd17c09f9/MIL/C04_Sets_and_Functions/solutions/Solutions_S01_Sets.lean#L10}{C04\_S01:10} & 3 & 1 & No & \textbf{0} & \textbf{Yes} & \textbf{0} & \textbf{Yes} \\
    \href{https://github.com/leanprover-community/mathematics_in_lean/blob/53e574e0cccd781578aa4f43e2f8633fd17c09f9/MIL/C04_Sets_and_Functions/solutions/Solutions_S01_Sets.lean#L15}{C04\_S01:15} & 7 & 2 & No & 2 & No & \textbf{0} & \textbf{Yes} \\
    \href{https://github.com/leanprover-community/mathematics_in_lean/blob/53e574e0cccd781578aa4f43e2f8633fd17c09f9/MIL/C04_Sets_and_Functions/solutions/Solutions_S01_Sets.lean#L28}{C04\_S01:28} & 5 & \textbf{0} & \textbf{Yes} & \textbf{0} & \textbf{Yes} & \textbf{0} & \textbf{Yes} \\
    \href{https://github.com/leanprover-community/mathematics_in_lean/blob/53e574e0cccd781578aa4f43e2f8633fd17c09f9/MIL/C04_Sets_and_Functions/solutions/Solutions_S01_Sets.lean#L35}{C04\_S01:35} & 3 & \textbf{0} & \textbf{Yes} & \textbf{0} & \textbf{Yes} & \textbf{0} & \textbf{Yes} \\
    \href{https://github.com/leanprover-community/mathematics_in_lean/blob/53e574e0cccd781578aa4f43e2f8633fd17c09f9/MIL/C04_Sets_and_Functions/solutions/Solutions_S01_Sets.lean#L40}{C04\_S01:40} & 14 & \textbf{0} & \textbf{Yes} & \textbf{0} & \textbf{Yes} & \textbf{0} & \textbf{Yes} \\
    \href{https://github.com/leanprover-community/mathematics_in_lean/blob/53e574e0cccd781578aa4f43e2f8633fd17c09f9/MIL/C04_Sets_and_Functions/solutions/Solutions_S01_Sets.lean#L56}{C04\_S01:56} & 21 & \textbf{0} & \textbf{Yes} & 18 & No & \textbf{0} & \textbf{Yes} \\
    \href{https://github.com/leanprover-community/mathematics_in_lean/blob/53e574e0cccd781578aa4f43e2f8633fd17c09f9/MIL/C04_Sets_and_Functions/solutions/Solutions_S01_Sets.lean#L79}{C04\_S01:79} & 7 & 7 & No & 5 & No & 5 & No \\
    \href{https://github.com/leanprover-community/mathematics_in_lean/blob/53e574e0cccd781578aa4f43e2f8633fd17c09f9/MIL/C04_Sets_and_Functions/solutions/Solutions_S01_Sets.lean#L97}{C04\_S01:97} & 4 & \textbf{0} & \textbf{Yes} & \textbf{0} & \textbf{Yes} & \textbf{0} & \textbf{Yes} \\
    \href{https://github.com/leanprover-community/mathematics_in_lean/blob/53e574e0cccd781578aa4f43e2f8633fd17c09f9/MIL/C04_Sets_and_Functions/solutions/Solutions_S01_Sets.lean#L103}{C04\_S01:103} & 2 & \textbf{0} & \textbf{Yes} & \textbf{0} & \textbf{Yes} & \textbf{0} & \textbf{Yes} \\
    \href{https://github.com/leanprover-community/mathematics_in_lean/blob/53e574e0cccd781578aa4f43e2f8633fd17c09f9/MIL/C04_Sets_and_Functions/solutions/Solutions_S01_Sets.lean#L118}{C04\_S01:118} & 19 & 12 & No & 18 & No & \textbf{0} & \textbf{Yes} \\
    \href{https://github.com/leanprover-community/mathematics_in_lean/blob/53e574e0cccd781578aa4f43e2f8633fd17c09f9/MIL/C04_Sets_and_Functions/solutions/Solutions_S01_Sets.lean#L142}{C04\_S01:142} & 6 & 4 & No & 4 & No & 4 & No \\
    \href{https://github.com/leanprover-community/mathematics_in_lean/blob/53e574e0cccd781578aa4f43e2f8633fd17c09f9/MIL/C04_Sets_and_Functions/solutions/Solutions_S01_Sets.lean#L142}{C04\_S01:142} & 6 & 4 & No & 4 & No & 4 & No \\
    \href{https://github.com/leanprover-community/mathematics_in_lean/blob/53e574e0cccd781578aa4f43e2f8633fd17c09f9/MIL/C04_Sets_and_Functions/solutions/Solutions_S02_Functions.lean#L16}{C04\_S02:16} & 8 & \textbf{0} & \textbf{Yes} & \textbf{0} & \textbf{Yes} & \textbf{0} & \textbf{Yes} \\
    \href{https://github.com/leanprover-community/mathematics_in_lean/blob/53e574e0cccd781578aa4f43e2f8633fd17c09f9/MIL/C04_Sets_and_Functions/solutions/Solutions_S02_Functions.lean#L26}{C04\_S02:26} & 3 & \textbf{0} & \textbf{Yes} & \textbf{0} & \textbf{Yes} & \textbf{0} & \textbf{Yes} \\
    \href{https://github.com/leanprover-community/mathematics_in_lean/blob/53e574e0cccd781578aa4f43e2f8633fd17c09f9/MIL/C04_Sets_and_Functions/solutions/Solutions_S02_Functions.lean#L31}{C04\_S02:31} & 2 & \textbf{0} & \textbf{Yes} & \textbf{0} & \textbf{Yes} & \textbf{0} & \textbf{Yes} \\
    \href{https://github.com/leanprover-community/mathematics_in_lean/blob/53e574e0cccd781578aa4f43e2f8633fd17c09f9/MIL/C04_Sets_and_Functions/solutions/Solutions_S02_Functions.lean#L35}{C04\_S02:35} & 8 & \textbf{0} & \textbf{Yes} & \textbf{0} & \textbf{Yes} & \textbf{0} & \textbf{Yes} \\
    \href{https://github.com/leanprover-community/mathematics_in_lean/blob/53e574e0cccd781578aa4f43e2f8633fd17c09f9/MIL/C04_Sets_and_Functions/solutions/Solutions_S02_Functions.lean#L45}{C04\_S02:45} & 2 & 1 & No & \textbf{0} & \textbf{Yes} & \textbf{0} & \textbf{Yes} \\
    \href{https://github.com/leanprover-community/mathematics_in_lean/blob/53e574e0cccd781578aa4f43e2f8633fd17c09f9/MIL/C04_Sets_and_Functions/solutions/Solutions_S02_Functions.lean#L49}{C04\_S02:49} & 1 & 1 & No & \textbf{0} & \textbf{Yes} & \textbf{0} & \textbf{Yes} \\
    \href{https://github.com/leanprover-community/mathematics_in_lean/blob/53e574e0cccd781578aa4f43e2f8633fd17c09f9/MIL/C04_Sets_and_Functions/solutions/Solutions_S02_Functions.lean#L52}{C04\_S02:52} & 1 & \textbf{0} & \textbf{Yes} & \textbf{0} & \textbf{Yes} & \textbf{0} & \textbf{Yes} \\
    \href{https://github.com/leanprover-community/mathematics_in_lean/blob/53e574e0cccd781578aa4f43e2f8633fd17c09f9/MIL/C04_Sets_and_Functions/solutions/Solutions_S02_Functions.lean#L55}{C04\_S02:55} & 4 & 2 & No & 2 & No & \textbf{0} & \textbf{Yes} \\
    \href{https://github.com/leanprover-community/mathematics_in_lean/blob/53e574e0cccd781578aa4f43e2f8633fd17c09f9/MIL/C04_Sets_and_Functions/solutions/Solutions_S02_Functions.lean#L61}{C04\_S02:61} & 7 & 2 & No & 2 & No & \textbf{0} & \textbf{Yes} \\
    \href{https://github.com/leanprover-community/mathematics_in_lean/blob/53e574e0cccd781578aa4f43e2f8633fd17c09f9/MIL/C04_Sets_and_Functions/solutions/Solutions_S02_Functions.lean#L70}{C04\_S02:70} & 7 & 1 & No & 5 & No & \textbf{0} & \textbf{Yes} \\
    \href{https://github.com/leanprover-community/mathematics_in_lean/blob/53e574e0cccd781578aa4f43e2f8633fd17c09f9/MIL/C04_Sets_and_Functions/solutions/Solutions_S02_Functions.lean#L84}{C04\_S02:84} & 5 & \textbf{0} & \textbf{Yes} & \textbf{0} & \textbf{Yes} & \textbf{0} & \textbf{Yes} \\
    \href{https://github.com/leanprover-community/mathematics_in_lean/blob/53e574e0cccd781578aa4f43e2f8633fd17c09f9/MIL/C04_Sets_and_Functions/solutions/Solutions_S02_Functions.lean#L91}{C04\_S02:91} & 2 & 1 & No & 2 & No & 1 & No \\
    \href{https://github.com/leanprover-community/mathematics_in_lean/blob/53e574e0cccd781578aa4f43e2f8633fd17c09f9/MIL/C04_Sets_and_Functions/solutions/Solutions_S02_Functions.lean#L95}{C04\_S02:95} & 2 & 1 & No & 2 & No & 1 & No \\
    \href{https://github.com/leanprover-community/mathematics_in_lean/blob/53e574e0cccd781578aa4f43e2f8633fd17c09f9/MIL/C04_Sets_and_Functions/solutions/Solutions_S02_Functions.lean#L99}{C04\_S02:99} & 4 & 1 & No & 2 & No & \textbf{0} & \textbf{Yes} \\
    \href{https://github.com/leanprover-community/mathematics_in_lean/blob/53e574e0cccd781578aa4f43e2f8633fd17c09f9/MIL/C04_Sets_and_Functions/solutions/Solutions_S02_Functions.lean#L107}{C04\_S02:107} & 6 & \textbf{0} & \textbf{Yes} & \textbf{0} & \textbf{Yes} & \textbf{0} & \textbf{Yes} \\
    \href{https://github.com/leanprover-community/mathematics_in_lean/blob/53e574e0cccd781578aa4f43e2f8633fd17c09f9/MIL/C04_Sets_and_Functions/solutions/Solutions_S02_Functions.lean#L115}{C04\_S02:115} & 4 & 1 & No & 1 & No & \textbf{0} & \textbf{Yes} \\
    \href{https://github.com/leanprover-community/mathematics_in_lean/blob/53e574e0cccd781578aa4f43e2f8633fd17c09f9/MIL/C04_Sets_and_Functions/solutions/Solutions_S02_Functions.lean#L121}{C04\_S02:121} & 11 & 8 & No & 8 & No & 6 & No \\
    \href{https://github.com/leanprover-community/mathematics_in_lean/blob/53e574e0cccd781578aa4f43e2f8633fd17c09f9/MIL/C04_Sets_and_Functions/solutions/Solutions_S02_Functions.lean#L134}{C04\_S02:134} & 2 & \textbf{0} & \textbf{Yes} & \textbf{0} & \textbf{Yes} & \textbf{0} & \textbf{Yes} \\
    \href{https://github.com/leanprover-community/mathematics_in_lean/blob/53e574e0cccd781578aa4f43e2f8633fd17c09f9/MIL/C04_Sets_and_Functions/solutions/Solutions_S02_Functions.lean#L138}{C04\_S02:138} & 2 & \textbf{0} & \textbf{Yes} & 1 & No & 1 & No \\
    \href{https://github.com/leanprover-community/mathematics_in_lean/blob/53e574e0cccd781578aa4f43e2f8633fd17c09f9/MIL/C04_Sets_and_Functions/solutions/Solutions_S02_Functions.lean#L148}{C04\_S02:148} & 3 & 1 & No & 3 & No & \textbf{0} & \textbf{Yes} \\
    \href{https://github.com/leanprover-community/mathematics_in_lean/blob/53e574e0cccd781578aa4f43e2f8633fd17c09f9/MIL/C04_Sets_and_Functions/solutions/Solutions_S02_Functions.lean#L157}{C04\_S02:157} & 4 & 1 & No & 2 & No & \textbf{0} & \textbf{Yes} \\
    \href{https://github.com/leanprover-community/mathematics_in_lean/blob/53e574e0cccd781578aa4f43e2f8633fd17c09f9/MIL/C04_Sets_and_Functions/solutions/Solutions_S02_Functions.lean#L167}{C04\_S02:167} & 10 & 5 & No & 4 & No & 4 & No \\
    \href{https://github.com/leanprover-community/mathematics_in_lean/blob/53e574e0cccd781578aa4f43e2f8633fd17c09f9/MIL/C04_Sets_and_Functions/solutions/Solutions_S02_Functions.lean#L179}{C04\_S02:179} & 8 & 7 & No & 8 & No & \textbf{0} & \textbf{Yes} \\
    \href{https://github.com/leanprover-community/mathematics_in_lean/blob/53e574e0cccd781578aa4f43e2f8633fd17c09f9/MIL/C04_Sets_and_Functions/solutions/Solutions_S02_Functions.lean#L209}{C04\_S02:209} & 7 & 6 & No & 6 & No & 2 & No \\
    \href{https://github.com/leanprover-community/mathematics_in_lean/blob/53e574e0cccd781578aa4f43e2f8633fd17c09f9/MIL/C04_Sets_and_Functions/solutions/Solutions_S02_Functions.lean#L221}{C04\_S02:221} & 7 & 4 & No & 3 & No & 3 & No \\
    \href{https://github.com/leanprover-community/mathematics_in_lean/blob/53e574e0cccd781578aa4f43e2f8633fd17c09f9/MIL/C04_Sets_and_Functions/solutions/Solutions_S02_Functions.lean#L239}{C04\_S02:239} & 10 & 8 & No & 9 & No & 8 & No \\
    \href{https://github.com/leanprover-community/mathematics_in_lean/blob/53e574e0cccd781578aa4f43e2f8633fd17c09f9/MIL/C04_Sets_and_Functions/solutions/Solutions_S03_The_Schroeder_Bernstein_Theorem.lean#L25}{C04\_S03:25} & 10 & 10 & No & 10 & No & 10 & No \\
    \href{https://github.com/leanprover-community/mathematics_in_lean/blob/53e574e0cccd781578aa4f43e2f8633fd17c09f9/MIL/C05_Elementary_Number_Theory/solutions/Solutions_S01_Irrational_Roots.lean#L9}{C05\_S01:9} & 22 & 16 & No & 19 & No & 16 & No \\
    \href{https://github.com/leanprover-community/mathematics_in_lean/blob/53e574e0cccd781578aa4f43e2f8633fd17c09f9/MIL/C05_Elementary_Number_Theory/solutions/Solutions_S01_Irrational_Roots.lean#L34}{C05\_S01:34} & 26 & 17 & No & 25 & No & 17 & No \\
    \href{https://github.com/leanprover-community/mathematics_in_lean/blob/53e574e0cccd781578aa4f43e2f8633fd17c09f9/MIL/C05_Elementary_Number_Theory/solutions/Solutions_S01_Irrational_Roots.lean#L76}{C05\_S01:76} & 10 & 9 & No & 8 & No & 6 & No \\
    \href{https://github.com/leanprover-community/mathematics_in_lean/blob/53e574e0cccd781578aa4f43e2f8633fd17c09f9/MIL/C05_Elementary_Number_Theory/solutions/Solutions_S01_Irrational_Roots.lean#L89}{C05\_S01:89} & 12 & 12 & No & 12 & No & 12 & No \\
    \href{https://github.com/leanprover-community/mathematics_in_lean/blob/53e574e0cccd781578aa4f43e2f8633fd17c09f9/MIL/C05_Elementary_Number_Theory/solutions/Solutions_S02_Induction_and_Recursion.lean#L8}{C05\_S02:8} & 9 & 7 & No & 7 & No & 7 & No \\
    \href{https://github.com/leanprover-community/mathematics_in_lean/blob/53e574e0cccd781578aa4f43e2f8633fd17c09f9/MIL/C05_Elementary_Number_Theory/solutions/Solutions_S02_Induction_and_Recursion.lean#L26}{C05\_S02:26} & 6 & 6 & No & 5 & No & 5 & No \\
    \href{https://github.com/leanprover-community/mathematics_in_lean/blob/53e574e0cccd781578aa4f43e2f8633fd17c09f9/MIL/C05_Elementary_Number_Theory/solutions/Solutions_S02_Induction_and_Recursion.lean#L67}{C05\_S02:67} & 4 & 3 & No & 1 & No & \textbf{0} & \textbf{Yes} \\
    \href{https://github.com/leanprover-community/mathematics_in_lean/blob/53e574e0cccd781578aa4f43e2f8633fd17c09f9/MIL/C05_Elementary_Number_Theory/solutions/Solutions_S02_Induction_and_Recursion.lean#L73}{C05\_S02:73} & 3 & 3 & No & 3 & No & \textbf{0} & \textbf{Yes} \\
    \href{https://github.com/leanprover-community/mathematics_in_lean/blob/53e574e0cccd781578aa4f43e2f8633fd17c09f9/MIL/C05_Elementary_Number_Theory/solutions/Solutions_S02_Induction_and_Recursion.lean#L78}{C05\_S02:78} & 4 & 1 & No & 1 & No & \textbf{0} & \textbf{Yes} \\
    \href{https://github.com/leanprover-community/mathematics_in_lean/blob/53e574e0cccd781578aa4f43e2f8633fd17c09f9/MIL/C05_Elementary_Number_Theory/solutions/Solutions_S02_Induction_and_Recursion.lean#L84}{C05\_S02:84} & 4 & 3 & No & 3 & No & 3 & No \\
    \href{https://github.com/leanprover-community/mathematics_in_lean/blob/53e574e0cccd781578aa4f43e2f8633fd17c09f9/MIL/C05_Elementary_Number_Theory/solutions/Solutions_S03_Infinitely_Many_Primes.lean#L33}{C05\_S03:33} & 19 & 16 & No & 18 & No & 15 & No \\
    \href{https://github.com/leanprover-community/mathematics_in_lean/blob/53e574e0cccd781578aa4f43e2f8633fd17c09f9/MIL/C05_Elementary_Number_Theory/solutions/Solutions_S03_Infinitely_Many_Primes.lean#L59}{C05\_S03:59} & 3 & \textbf{0} & \textbf{Yes} & \textbf{0} & \textbf{Yes} & \textbf{0} & \textbf{Yes} \\
    \href{https://github.com/leanprover-community/mathematics_in_lean/blob/53e574e0cccd781578aa4f43e2f8633fd17c09f9/MIL/C05_Elementary_Number_Theory/solutions/Solutions_S03_Infinitely_Many_Primes.lean#L69}{C05\_S03:69} & 3 & \textbf{0} & \textbf{Yes} & 1 & No & \textbf{0} & \textbf{Yes} \\
    \href{https://github.com/leanprover-community/mathematics_in_lean/blob/53e574e0cccd781578aa4f43e2f8633fd17c09f9/MIL/C05_Elementary_Number_Theory/solutions/Solutions_S03_Infinitely_Many_Primes.lean#L81}{C05\_S03:81} & 3 & 1 & No & 2 & No & 1 & No \\
    \href{https://github.com/leanprover-community/mathematics_in_lean/blob/53e574e0cccd781578aa4f43e2f8633fd17c09f9/MIL/C05_Elementary_Number_Theory/solutions/Solutions_S03_Infinitely_Many_Primes.lean#L88}{C05\_S03:88} & 11 & 8 & No & 8 & No & 8 & No \\
    \href{https://github.com/leanprover-community/mathematics_in_lean/blob/53e574e0cccd781578aa4f43e2f8633fd17c09f9/MIL/C05_Elementary_Number_Theory/solutions/Solutions_S03_Infinitely_Many_Primes.lean#L102}{C05\_S03:102} & 26 & 25 & No & 22 & No & 22 & No \\
    \href{https://github.com/leanprover-community/mathematics_in_lean/blob/53e574e0cccd781578aa4f43e2f8633fd17c09f9/MIL/C05_Elementary_Number_Theory/solutions/Solutions_S03_Infinitely_Many_Primes.lean#L160}{C05\_S03:160} & 3 & 3 & No & 3 & No & \textbf{0} & \textbf{Yes} \\
    \href{https://github.com/leanprover-community/mathematics_in_lean/blob/53e574e0cccd781578aa4f43e2f8633fd17c09f9/MIL/C05_Elementary_Number_Theory/solutions/Solutions_S03_Infinitely_Many_Primes.lean#L165}{C05\_S03:165} & 27 & 27 & No & 27 & No & 26 & No \\
    \href{https://github.com/leanprover-community/mathematics_in_lean/blob/53e574e0cccd781578aa4f43e2f8633fd17c09f9/MIL/C06_Structures/solutions/Solutions_S01_Structures.lean#L19}{C06\_S01:19} & 1 & 1 & No & 1 & No & \textbf{0} & \textbf{Yes} \\
    \href{https://github.com/leanprover-community/mathematics_in_lean/blob/53e574e0cccd781578aa4f43e2f8633fd17c09f9/MIL/C06_Structures/solutions/Solutions_S01_Structures.lean#L25}{C06\_S01:25} & 1 & 1 & No & 1 & No & 1 & No \\
    \href{https://github.com/leanprover-community/mathematics_in_lean/blob/53e574e0cccd781578aa4f43e2f8633fd17c09f9/MIL/C06_Structures/solutions/Solutions_S03_Building_the_Gaussian_Integers.lean#L196}{C06\_S03:196} & 2 & 2 & No & \textbf{0} & \textbf{Yes} & \textbf{0} & \textbf{Yes} \\
    \href{https://github.com/leanprover-community/mathematics_in_lean/blob/53e574e0cccd781578aa4f43e2f8633fd17c09f9/MIL/C06_Structures/solutions/Solutions_S03_Building_the_Gaussian_Integers.lean#L200}{C06\_S03:200} & 2 & 1 & No & 1 & No & 1 & No \\
    \href{https://github.com/leanprover-community/mathematics_in_lean/blob/53e574e0cccd781578aa4f43e2f8633fd17c09f9/MIL/C06_Structures/solutions/Solutions_S03_Building_the_Gaussian_Integers.lean#L204}{C06\_S03:204} & 2 & 1 & No & 1 & No & 1 & No \\
    \href{https://github.com/leanprover-community/mathematics_in_lean/blob/53e574e0cccd781578aa4f43e2f8633fd17c09f9/MIL/C06_Structures/solutions/Solutions_S03_Building_the_Gaussian_Integers.lean#L208}{C06\_S03:208} & 2 & 2 & No & 1 & No & \textbf{0} & \textbf{Yes} \\
    \href{https://github.com/leanprover-community/mathematics_in_lean/blob/53e574e0cccd781578aa4f43e2f8633fd17c09f9/MIL/C07_Hierarchies/solutions/Solutions_S01_Basics.lean#L181}{C07\_S01:181} & 1 & 1 & No & 1 & No & \textbf{0} & \textbf{Yes} \\
    \href{https://github.com/leanprover-community/mathematics_in_lean/blob/53e574e0cccd781578aa4f43e2f8633fd17c09f9/MIL/C07_Hierarchies/solutions/Solutions_S01_Basics.lean#L185}{C07\_S01:185} & 1 & 1 & No & 1 & No & 1 & No \\
    \href{https://github.com/leanprover-community/mathematics_in_lean/blob/53e574e0cccd781578aa4f43e2f8633fd17c09f9/MIL/C07_Hierarchies/solutions/Solutions_S01_Basics.lean#L189}{C07\_S01:189} & 1 & 1 & No & 1 & No & 1 & No \\
    \href{https://github.com/leanprover-community/mathematics_in_lean/blob/53e574e0cccd781578aa4f43e2f8633fd17c09f9/MIL/C07_Hierarchies/solutions/Solutions_S01_Basics.lean#L206}{C07\_S01:206} & 3 & 3 & No & 3 & No & 3 & No \\
    \href{https://github.com/leanprover-community/mathematics_in_lean/blob/53e574e0cccd781578aa4f43e2f8633fd17c09f9/MIL/C07_Hierarchies/solutions/Solutions_S03_Subobjects.lean#L94}{C07\_S03:94} & 3 & 3 & No & 3 & No & 3 & No \\
    \href{https://github.com/leanprover-community/mathematics_in_lean/blob/53e574e0cccd781578aa4f43e2f8633fd17c09f9/MIL/C07_Hierarchies/solutions/Solutions_S03_Subobjects.lean#L110}{C07\_S03:110} & 10 & 10 & No & 10 & No & 8 & No \\
    \href{https://github.com/leanprover-community/mathematics_in_lean/blob/53e574e0cccd781578aa4f43e2f8633fd17c09f9/MIL/C08_Groups_and_Rings/solutions/Solutions_S01_Groups.lean#L8}{C08\_S01:8} & 13 & 4 & No & 6 & No & 3 & No \\
    \href{https://github.com/leanprover-community/mathematics_in_lean/blob/53e574e0cccd781578aa4f43e2f8633fd17c09f9/MIL/C08_Groups_and_Rings/solutions/Solutions_S01_Groups.lean#L32}{C08\_S01:32} & 3 & 1 & No & \textbf{0} & \textbf{Yes} & \textbf{0} & \textbf{Yes} \\
    \href{https://github.com/leanprover-community/mathematics_in_lean/blob/53e574e0cccd781578aa4f43e2f8633fd17c09f9/MIL/C08_Groups_and_Rings/solutions/Solutions_S01_Groups.lean#L37}{C08\_S01:37} & 3 & 1 & No & \textbf{0} & \textbf{Yes} & \textbf{0} & \textbf{Yes} \\
    \href{https://github.com/leanprover-community/mathematics_in_lean/blob/53e574e0cccd781578aa4f43e2f8633fd17c09f9/MIL/C08_Groups_and_Rings/solutions/Solutions_S02_Rings.lean#L33}{C08\_S02:33} & 1 & 1 & No & 1 & No & 1 & No \\
    \href{https://github.com/leanprover-community/mathematics_in_lean/blob/53e574e0cccd781578aa4f43e2f8633fd17c09f9/MIL/C09_Topology/solutions/Solutions_S01_Filters.lean#L63}{C09\_S01:63} & 3 & 3 & No & 3 & No & 1 & No \\
    \href{https://github.com/leanprover-community/mathematics_in_lean/blob/53e574e0cccd781578aa4f43e2f8633fd17c09f9/MIL/C09_Topology/solutions/Solutions_S02_Metric_Spaces.lean#L92}{C09\_S02:92} & 7 & 6 & No & 6 & No & \textbf{0} & \textbf{Yes} \\
    \href{https://github.com/leanprover-community/mathematics_in_lean/blob/53e574e0cccd781578aa4f43e2f8633fd17c09f9/MIL/C09_Topology/solutions/Solutions_S03_Topological_Spaces.lean#L55}{C09\_S03:55} & 4 & 3 & No & 2 & No & \textbf{0} & \textbf{Yes} \\
    \href{https://github.com/leanprover-community/mathematics_in_lean/blob/53e574e0cccd781578aa4f43e2f8633fd17c09f9/MIL/C09_Topology/solutions/Solutions_S03_Topological_Spaces.lean#L112}{C09\_S03:112} & 1 & 1 & No & 1 & No & 1 & No \\
    \href{https://github.com/leanprover-community/mathematics_in_lean/blob/53e574e0cccd781578aa4f43e2f8633fd17c09f9/MIL/C09_Topology/solutions/Solutions_S03_Topological_Spaces.lean#L118}{C09\_S03:118} & 19 & 19 & No & 19 & No & 18 & No
    
\label{table:results_details}
\end{longtable}

%% file: sections/impact.tex
\section{Impacts \& Future Works}
\label{sec:impacts}

In this paper, we have presented {\lc}, a general open-source framework for running LLM inference in Lean.
This solves a historical challenge in Lean of using the power of \textit{neural} networks to aid development in the \textit{symbolic} system of the Lean proof assistant.
Therefore, {\lc} opens up a wide range of opportunities for neuro-symbolic reasoning in Lean.
In this section, we highlight some immediate impacts {\lc} may have in the field, in the near future.

\begin{enumerate}

    \item \textbf{LLM-Powered Proof Automation for Lean Users}: Lean users can now use our proof automation tools easily in Lean, to get tactic suggestions, proof completions, and premise selections, which accelerate the theorem proving process in Lean. 
    We are glad to have already heard many such reports from the Lean community, and greatly appreciate people's feedback.
    
    \item \textbf{Building More Proof Automation}: In this work, we build proof automation tools for three highly important applications, yet there are certainly more such tools that can be built to aid theorem proving in Lean. 
    Example include autoformalization, proof repair, LLM-guided proof search algorithms, and more. 
    We are happy to have already seen such efforts being proposed and added to {\lc}, including proof progress prediction \citep{huang2025leanprogressguidingsearchneural} and lifelong learning \citep{kumarappan2024leanagentlifelonglearningformal}.
    
    \item \textbf{More Immediate Synchronization within the Field}: With {\lc} being a general framework that supports user-brought models, once a new state-of-the-art machine learning model appears on the \textit{neural} side of the field, it can be immediately added via {\lc} to the \textit{symbolic} system of Lean.
    On the other hand, {\lc} can also help AI developers test their models in the real Lean enviroment. 
    In other words, {\lc} as a neuro-symbolic framework can serve as a bridge between the \textit{neural} and the \textit{symbolic} sides of the field, enabling more effective synchronization.
    
    \item \textbf{Dynamic Copilots}: While {\lc} supports efficient inference of LLMs, there are times when (some light) learning is desired.
    Especially in fast-developing formalization projects, new premises can appear frequently.
    If the knowledge base of the LLM behind the proof automation tools is static, it would require the LLM to be updated once in a while. 
    Otherwise it soon becomes less helpful.
    There can be many ways to enable LLMs to learn dynamically from a growing Lean project, such as in-context learning \citep{dong2024surveyincontextlearning, min2022metaicllearninglearncontext} (which then requires satisfactory long-context abilities) or efficient ML \citep{zhao2024galorememoryefficientllmtraining, efficient_ml} (so that not only inference but also training can happen natively in Lean).
    Advances in those approaches can help make {\lc} dynamic and consistently helpful.
    
\end{enumerate}